\pdfoutput=1

\documentclass[journal]{IEEEtran}
\ifCLASSINFOpdf
  \usepackage[pdftex]{graphicx}
   \graphicspath{{../figures/}}
  \DeclareGraphicsExtensions{.pdf,.jpeg,.png}
\else
  \usepackage[dvips]{graphicx}
\fi
\usepackage{amsmath}
\usepackage{algorithmic}

\usepackage{array}
\usepackage{url}

\usepackage[caption=false]{subfig}
\usepackage[capitalise]{cleveref}
\usepackage{wrapfig}

\usepackage{color}
\usepackage[switch]{lineno} 
\usepackage[hidelinks]{hyperref}
\usepackage{xcolor}
\usepackage{amssymb}
\usepackage{multirow}
\usepackage{mathtools}
\usepackage{booktabs}
\usepackage[normalem]{ulem}

\usepackage{soul}

\newcommand{\rev}[1]{}
\renewcommand{\rev}[1]{{\color{red}{#1}}}

\newcommand{\btodo}[1]{}
\renewcommand{\btodo}[1]{{\color{blue} bernhard: {#1}}}

\usepackage{tabularx}

\begin{document}
%
\title{PVR: Patch-to-Volume Reconstruction \\for Large Area Motion Correction of Fetal MRI   
}
%
%
%

\author{ 
	Amir Alansary, Bernhard Kainz$^*$, Martin Rajchl, Maria Murgasova,  Mellisa Damodaram, David F.A. Lloyd, Alice Davidson, Steven G. McDonagh, 
	Mary Rutherford, Joseph V. Hajnal, and Daniel Rueckert
\vspace{-5pt}
\thanks{A. Alansary, B. Kainz, M. Rajchl, SG. McDonagh, and D. Rueckert are with Biomedical Image Analysis Group, Department of Computing, Imperial College London, UK 
	}
\thanks{M. Murgasova, M. Damodaram, DFA. Lloyd, A. Davidson, M. Rutherford, and JV. Hajnal  are with the Department of Biomedical Engineering King's College London, UK 
	}
}%
%
%

\markboth{Submitted to IEEE TRANSACTIONS ON MEDICAL IMAGING}%
{Shell \MakeLowercase{\textit{et al.}}: Bare Demo of IEEEtran.cls for IEEE Journals}
%

\maketitle

\begin{abstract}
In this paper we present a novel method for the correction of motion artifacts that are present in fetal Magnetic Resonance Imaging (MRI) scans of the whole uterus. Contrary to current slice-to-volume registration (SVR) methods, requiring an inflexible anatomical enclosure of a \emph{single} investigated organ, the proposed patch-to-volume reconstruction (PVR) approach is able to reconstruct a large field of view of non-rigidly deforming structures. It relaxes rigid motion assumptions by introducing a specific amount of redundant information that is exploited with parallelized patch-wise optimization, super-resolution, and automatic outlier rejection. We further describe and provide an efficient parallel implementation of PVR allowing its execution within reasonable time on commercially available graphics processing units (GPU), enabling its use in the clinical practice. We evaluate PVR's computational overhead compared to standard methods and observe improved reconstruction accuracy in presence of affine motion artifacts of approximately 30\% compared to conventional SVR in synthetic experiments. 
Furthermore, we have evaluated our method qualitatively and quantitatively on real fetal MRI data subject to maternal breathing and sudden fetal movements. We evaluate peak-signal-to-noise ratio (PSNR), structural similarity index (SSIM), and cross correlation (CC) with respect to the originally acquired data and provide a method for visual inspection of reconstruction uncertainty. With these experiments we demonstrate successful application of PVR motion compensation to the whole uterus, the human fetus, and the human placenta.

\end{abstract}


\IEEEpeerreviewmaketitle

\section{Introduction}
\IEEEPARstart{T}{he} recent advent of single shot fast spin echo (ssFSE) T2-weighted sequences has enabled Magnetic Resonance Imaging (MRI) to play an essential role in fetal diagnosis~\cite{ertl2002fetal} and research. In particular, cases for which ultrasound (US) fails to acquire conclusive image data benefit from fetal MRI~\cite{garel2008imaging}.
Fetal MRI is able to distinguish individual fetal structures such as brain, lung, kidney and liver, as well as pregnancy structures such as the placenta, umbilical cord and amniotic sac~\cite{frates2004fetal}. It provides improved visualization and structural information of the fetal anatomy, which helps to study abnormalities during pregnancy such as neuro-developmental disorders~\cite{woodward2006neonatal}, placental pathologies~\cite{linduska2009placental}, fetuses with congenital lung masses~\cite{zamora2014fetal}, and conjoined twins~\cite{pierro2015classification}. MRI is considered to be safe after the first trimester~\cite{frates2004fetal} for 1.5T \cite{patenaude2014use} and 3T~\cite{cannie2015potential} without the use of contrast agents, which may have teratogenic effects. Furthermore, this technology paves the way for researchers and clinicians to analyze correlations between childhood development and prenatal abnormalities. 

\begin{figure}[htbp]
\centering
	\null\hfill
	{%
		\includegraphics[width=1.0\columnwidth, trim = 30 0 0 0, clip]{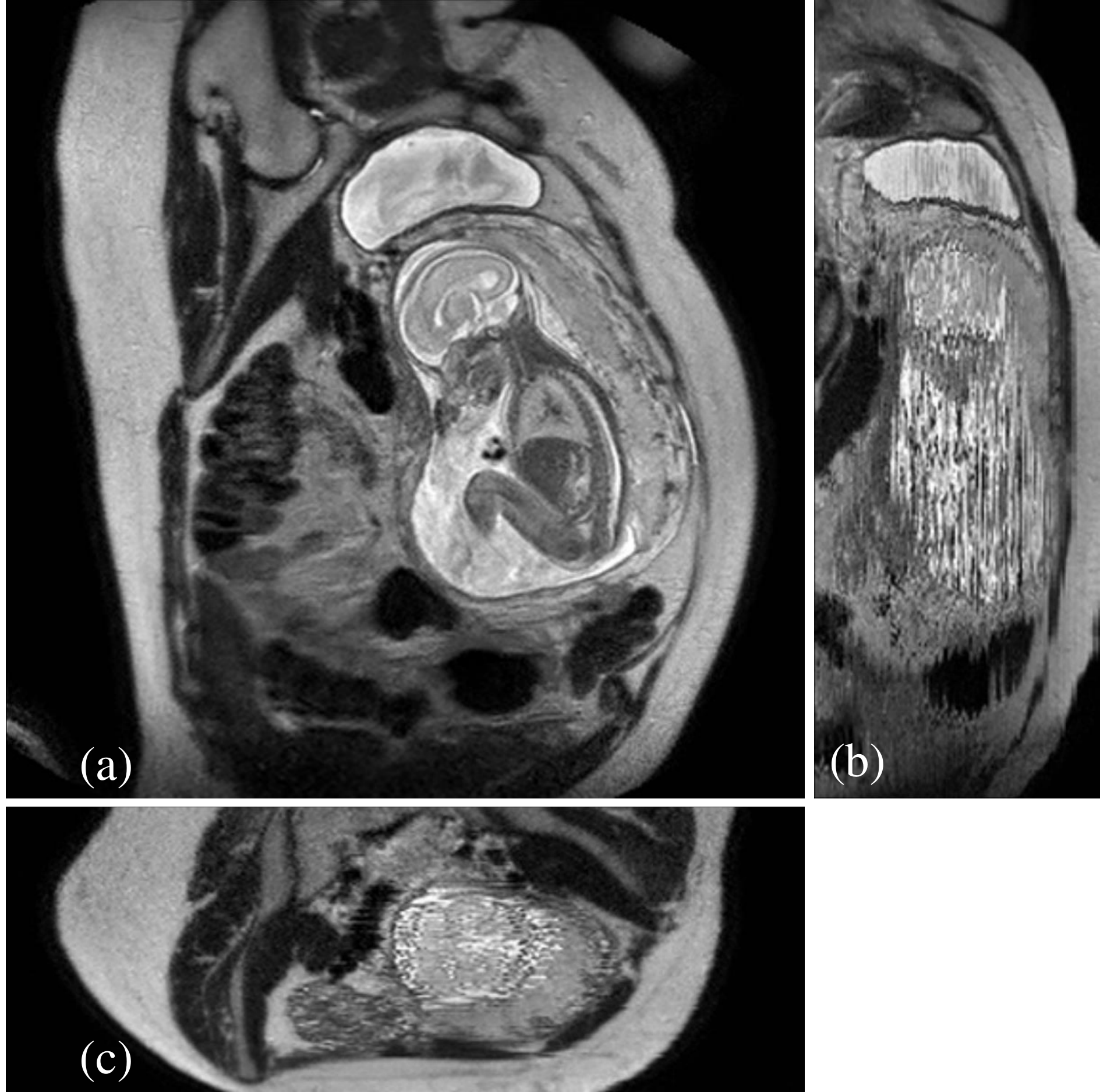}
	}
	\hfill\null
	\caption{Three view-planes for raw 3D data acquired through stacks of ssFSE images covering the whole uterus. The transverse (a) is the \emph{in-plane} view, \emph{i.e.}, native 2D slice scan orientation. Motion causes streaky artifacts for multi-planar reconstructions (MPR) in orthogonal views (b) and (c) caused by both maternal and fetal movements between the acquisition of individual slices.}
\label{fig:uterus}
\end{figure}

During image acquisition the fetus is not sedated and moves freely as well as the mother breathes normally. As a result, movements are likely to corrupt the scans, hiding pathology and causing overlap between different anatomical regions.
In order to limit these artifacts, fast scanning sequences such as ssFSE~\cite{Levine2004} allow for the rapid acquisition of single slices at high \emph{in-plane} resolution in a large field of view and good tissue contrast of the uterus. However, when acquiring a 3D volume through a stack of slices, inter-slice artifacts in the \emph{out-of-plane} views are highly likely. Consequently, this  restricts reliable diagnostics to individual slices in the current clinical practice. Fig.~\ref{fig:uterus} depicts a typical example of motion related artifacts in a fetal single-shot fast spin echo (ssFSE) scan. The observed motion (\emph{c.f.} Fig. \ref{fig:uterus} b \& c) is of unpredictable nature and consists of a combination of maternal respiration movements, fetal movements and bowel movements. 

\begin{figure}[htb]
	\centering
	\includegraphics[width=1.0\columnwidth]{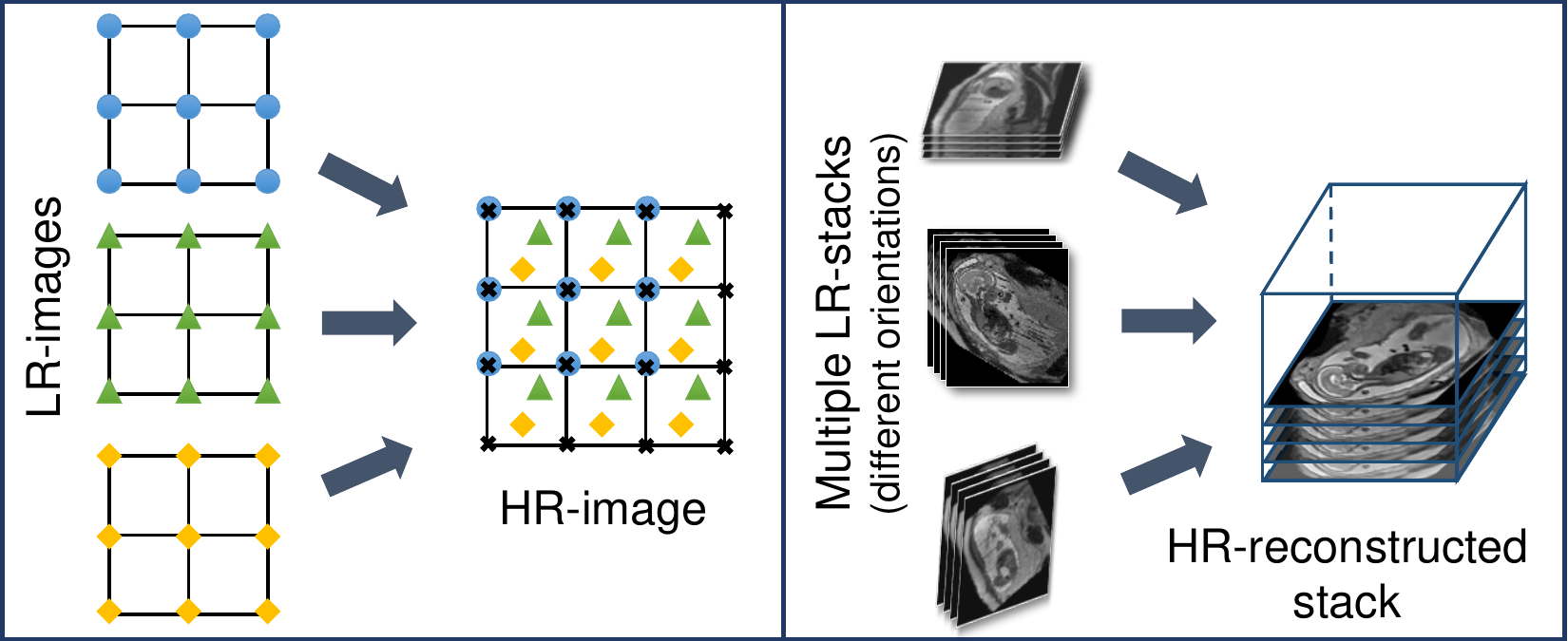}
	\caption[superresolution]{Illustration of the basic ideas behind reconstruction~\cite{park2003super}: A simplified example of a 2D $4$x$4$ HR grid sampling from a 2D $3$x$3$ LR grids (left) and a practical example of 3D fetal MRI using multiple overlapping stacks of slices, by reconstructing a 3D HR image with an isotropic voxel size from LR images with anisotropic voxel size.}
	\label{fig:superresolution}
\end{figure}

\noindent\textbf{Slice-to-volume registration (SVR):} SVR combined with super-resolution image reconstruction techniques~\cite{park2003super} can be applied to compensate motion between single slices by reconstructing a high-resolution (HR) image from multiple, overlapping low-resolution (LR) images, as shown in Fig.~\ref{fig:superresolution}. To provide a sufficiently high number of samples for such an approach, multiple stacks of 2D-slices need to be acquired, ideally in orthogonal orientations.
A simple LR $\rightarrow$ HR reconstruction model~\cite{park2003super} can be formalized as:
\begin{equation}
	x_i = W_{i}y + n_i \quad \textrm{for} \quad 1 \leq{i} \leq{N},
\label{eqn:reconstruction}
\end{equation}
\noindent where $x_i$ denotes the $i$-th LR image of total $N$ images, and $y$ is the HR image. The matrix $W_i$ combines motion, sub-sampling and degradation effects: $W_{i}=DBT_{i}$, where $D$ is the sub-sampling matrix, $B$ is the blurring matrix, and $T_i$ is the transformation matrix of observation $i$. The noise of observation $i$ is represented by $n_i$. Any LR image can be considered as a down-sampled, motion corrupted, blurred, and noisy version of the HR image. 
The resulting reconstruction problem can be divided into two main parts: motion correction (estimating $W_i$) and super-resolution reconstruction (estimating $y$). Image registration can be used for estimating motion, interpolation for obtaining a uniformly spaced HR image, and regularized super resolution with automatic outlier rejection for removing blur and noise.
Volumetric fetal MR image reconstruction is more challenging than typical image reconstruction problems due to unconstrained random motion during slice acquisitions. Slice misalignments can lead to a loss of spatial coherence and typically present with anisotropic voxel sizes and intensity inhomogeneities. 

\noindent\textbf{Practical limitations:} SVR methods have been successfully used to address these problems in fetal MR and are typically applied to small regions and organs with rigid body characteristics that are identified by manual, labor intensive~\cite{rousseau2006registration,kim2010intersection} annotations or less precise, automated segmentations~\cite{keraudren2014automated}. Such approaches are prohibitive to whole body and uterus reconstruction because of the assumption of rigid motion in the 2D to 3D registration step of SVR. As a result, different areas in each slice that are likely to move in different directions will break this assumption, \emph{e.g.}, the head and thorax. Further, an extension of 2D-3D registration to include non-rigid deformations is only well defined with each slice and not well-constrained in 3D. Current SVR approaches will fail in presence of non-rigid deformations and unpredictable organ shapes. This restricts the application of SVR to regions that are manually or automatically annotated. 
Thus, most of the previous SVR methods have been limited to the fetal brain as the main region of interest for fetal reconstruction due to the high incidence of neuro-developmental disability in premature infants. Only recently, \cite{kainz2014motion,Lloyd2016} proposed a motion corrected 3D reconstruction of fetal thoracic structures from prenatal MRI. 
Moreover, SVR is computationally expensive due to the exponential increase of computations with the size of the target area. This leads to prohibitive post-processing times in the clinical practice. Parallelized implementations~\cite{kainz2015fast} can address run-time problems, however, methodologically the SVR is still restricted to small, rigid body areas.

\noindent\textbf{Reconstruction of large-scale anatomy:} MRI has also been shown to be very useful for the evaluation of the whole uterus and structures like the placenta. During both, normal and high-risk pregnancies, the whole uterine appearance and the condition of the placenta are considered to be an indicator for fetal health after birth~\cite{routledge2015can}. Placental functions affect the birth weight as it controls the transmission of nutrients from the maternal to the fetal circulation~\cite{salafia2009allometric}.
However, the whole fetal body and secondary uterine parts can be inherently inconsistent. Different fetal body parts can move independently from the uterus. This makes the application of SVR and 2D-3D registration to the full uterus impossible in the presence of fetal motion and maternal respiration.

Besides, multiple births is a case where classical SVR pipelines based on preprocessing steps to identify consistent rigid regions will likely fail. The presence of multiple instances of the same fetal structure is usually not considered in previous methods.
Therefore, a fully automatic motion correction method for the whole uterus, as it is presented in this paper, is very desirable and will enable the application of standard 3D image analysis techniques, \emph{e.g.}, \cite{keraudren2015automated,alansary2016fast}.

\subsection{Related Work}
\begin{figure*}[htb]
	\centering
	\includegraphics[width=1.0\textwidth]{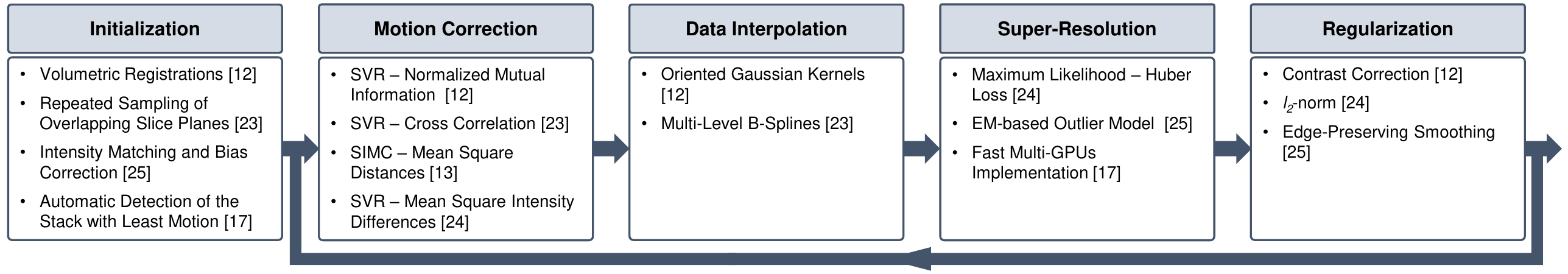}
	\caption[Related Work]{Overview of the required modules of state-of-the-art SVR methods and main components introduced by previous work.}
	\label{fig:related_work}
\end{figure*}

Most motion compensation approaches for fetal MRI are based on SVR techniques that aim to obtain a motion-free and high resolution volume of a fetal target region. Registration of individual 2D slices with a higher resolution 3D volume~\cite{fei2003slice} is the core approach of these algorithms. SVR methods assume that all acquired image stacks are centered at a specific fetal organ (\emph{e.g.}, brain, thorax) and cover three orthogonal image directions. Fig. \ref{fig:related_work} shows the core elements of SVR and the contribution of previous frameworks from the literature. 

The first SVR-based reconstruction framework for fetal imaging was introduced by Rousseau et al.~\cite{rousseau2006registration}. It includes steps to correct 2D slice misalignments, intensity inhomogeneity distortions, and reconstructs an isotropic HR fetal MRI brain image from sets of LR images. Motion correction is done by applying a global 3D rigid alignment between the LR images using one image as a reference to define the global coordinate system. Then every slice of these images is aligned to the initial reconstructed HR volume. Normalized mutual information is maximized using the gradient ascent method for both registration steps. 
A narrow Gaussian kernel is applied as a point spread function (PSF) for volume reconstruction and empty voxels are filled using the mean of the surrounding voxels. The image contrast is corrected using one LR image as a reference. 

Jiang et al.~\cite{jiang2007mri} introduces the acquisition of many thin slices to provide sufficient sampling of the region of interest. Cross correlation is used as a cost function for the SVR steps assuming that the data have consistent contrast properties. After that, multilevel B-splines are applied to the volumetric reconstruction for data interpolation, which has the advantage of reducing blurring of the reconstructed image supported by including the thin slices. 

Kim et al.~\cite{kim2010intersection} propose a method for slice intersection motion correction (SIMC) of multi-slice MRI for 3D fetal brain image formation. The method is based on slice-to-slice registration using spatially weighted mean square intensity differences (MSD) of the signal between slices as an energy, assuming that the MRI contrasts are identical. Maternal tissues are excluded from the energy computations using a windowing function of a parametric ellipsoid model. Similar to \cite{rousseau2006registration}, temporally adjacent slices are grouped together then divided into half iteratively. The splitting process is performed using discrete cosine basis functions. 

Gholipour et al.~\cite{gholipour2010robust} were the first to introduce a mathematical model for super-resolution (SR) volume reconstruction from slice acquisitions of fetal brains. The main difference to previous reconstruction methods is that it includes knowledge of the slice acquisition model and the SR reconstruction is performed based on maximum likelihood and a robust M-estimation minimization for an error norm function. A regularization term is also added to the cost function in order to enforce a solution when the number of acquired samples is not high enough for solving the reconstruction problem.

Murgasova et al.~\cite{murgasova2012reconstruction} were able to reconstruct the fetal brain using intensity matching and complete outlier removal. The main steps of their reconstruction method are: ({\it{i}}) 3D registration of the acquired stacks using a template stack; ({\it{ii}}) extracting region of interest (the fetal head) from all the stacks; ({\it{iii}}) intensity matching and bias correction between the slices based on an EM framework, where the differential bias fields and slice-dependent scaling factors are estimated during the reconstruction; ({\it{iv}}) motion correction using~\cite{rousseau2006registration} based on the normalized cross correlation as a similarity measure and an approximated 3D Gaussian PSF similar to \cite{jiang2007mri}. A posterior probability is used to define the inlier and outlier voxels within the EM framework in order to remove the motion-corrupted artifacts and misaligned data. Blurring in reconstructed images is reduced by integrating edge-preserving regularization based on anisotropic diffusion within the SR reconstruction framework.

Kainz et al. \cite{kainz2015fast} developed a fast multi-GPU accelerated implementation for the method presented in~\cite{murgasova2012reconstruction}, which is based on 2D-3D registration, SR with automatic outlier rejection and an optional intensity bias correction. They extended the reconstruction framework by automatically selecting the stack with least motion as the reference stack and using a fully flexible and accurate PSF instead of approximated functions. Using a multi-GPU framework enabled the SR reconstruction process to be approximately five to ten times faster than using a multi-CPU framework.

\subsection{Contributions}
In this paper we propose and evaluate a new paradigm for motion correction based on SVR and flexible subdivision of the input space into overlapping, highly redundant and partly rigid image patches~\cite{kainz2015flexible}, thus solving the motion compensation problem for \uline{large field of view reconstructions}. We split the input into small overlapping areas and find these, which contain rigid components. This allows to iteratively learn their consistency compared to a global reconstruction optimization volume. Corrupted and inconsistent data is \uline{automatically identified and excluded using robust statistics}.
 The proposed approach facilitates the \uline{automatic reconstruction} of whole collections of motion corrupted stacks \uline{without the need of corresponding image segmentations}. By treating rigid image patches as piecewise constant segments of organs further allows limited correction of non-rigid tissue motion. The presented patch-to-volume reconstruction (PVR) method finds rigidly connected areas automatically, which can be used as segmentation prior for further refinement using conventional SVR in small regions of interest. In contrast to \cite{kainz2015fast}, we further introduce a multi-scale patch definition approach and thoroughly evaluate the reconstruction quality of the whole uterus including the fetal brain and placenta. We test the breaking points of SVR and variations of PVR on synthetically motion corrupted brain phantom data. The presented approach is the \uline{only currently available method} that is able to reconstruct fetal organs and detailed 3D volumes of secondary, non-rigidly moving  structures such as the placenta. 

\section{Method}
SVR-based motion compensation methods make use of the assumption that rigid regions, \emph{e.g.}, brain and thorax, of 2D input slices deforms rigidly, where a global 3D volume is reconstructed by iteratively registering these 2D input slices. We propose to increase the granularity of the input data by using 2D data \emph{patches} of arbitrary shape instead of whole slices for SVR reconstruction. We explore  square patches and dilated superpixels~\cite{achanta2012slic} for the definition of the patch shape. 
Superpixels provide a method to define semantically meaningful regions while reducing the required data redundancy and computational overhead. 

\begin{figure*}[!htbp]
	\centering
		\includegraphics[width=0.9\textwidth]{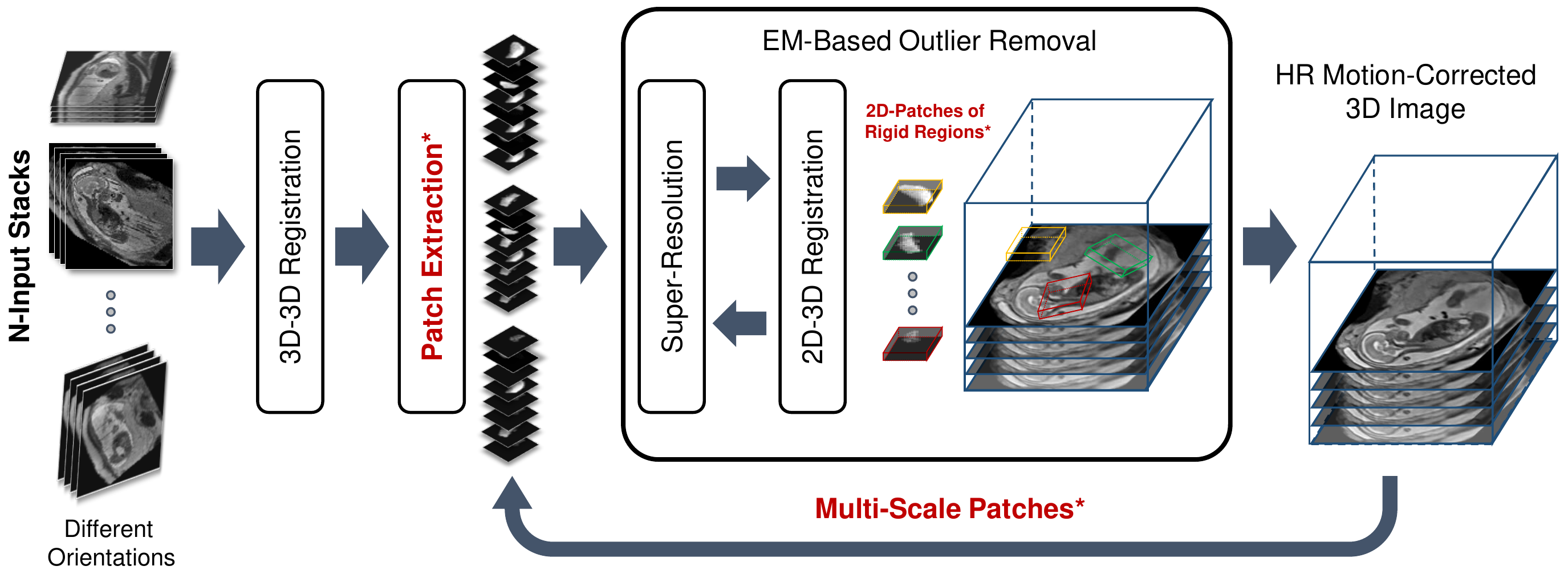}
	\caption[overview]{A schematic and modular overview of the proposed patch-to-volume reconstruction (PVR) framework.
	The key parts are 3D-3D registration, patch extraction, 2D-3D registration, super-resolution, and  EM-based outlier removal. 
	Core contributions of PVR are written in red and marked with asterisk.}
	\label{fig:overview}
\end{figure*}

PVR relies on the fact that certain regions of the scanned anatomy are rigid and can be reconstructed with SVR super-resolution algorithms. However, unlike SVR, it is fully automatic and provides a full field of view reconstruction. Data consistency is obtained by oversampling a region of interest at different scan orientations. Robust statistics can be used to identify mis-registered or heavily corrupted data~\cite{gholipour2010robust,murgasova2012reconstruction}. Fig. \ref{fig:overview} depicts a schematic overview of the proposed PVR framework.

\textbf{Input Data \& Initialization: } 
A template stack is either randomly or automatically chosen from available input stacks by detecting the stack with the least motion artifacts~\cite{kainz2015fast}. Global intensity matching is applied to normalize intensity values of all input images followed by global 3D-3D alignments to spatially initialize the reconstruction target. Input data can be represented as stacks of 2D images (patches) consisting of
\begin{eqnarray} 
    Y = \{y_s | s \in S\},
	\label{eq:patch}
\end{eqnarray}
where $y_s$ is a patch of arbitrary 2D-shape and indexed by the location $s$. $S$ is the set of all locations in all $p$ stacks, $S = \{s_1, s_2, ... s_M \}$, and $M$ is total number of patches.

\textbf{Patch Extraction:} 
\begin{figure}[h!]
	\centering
	\includegraphics[width=1.0\columnwidth]{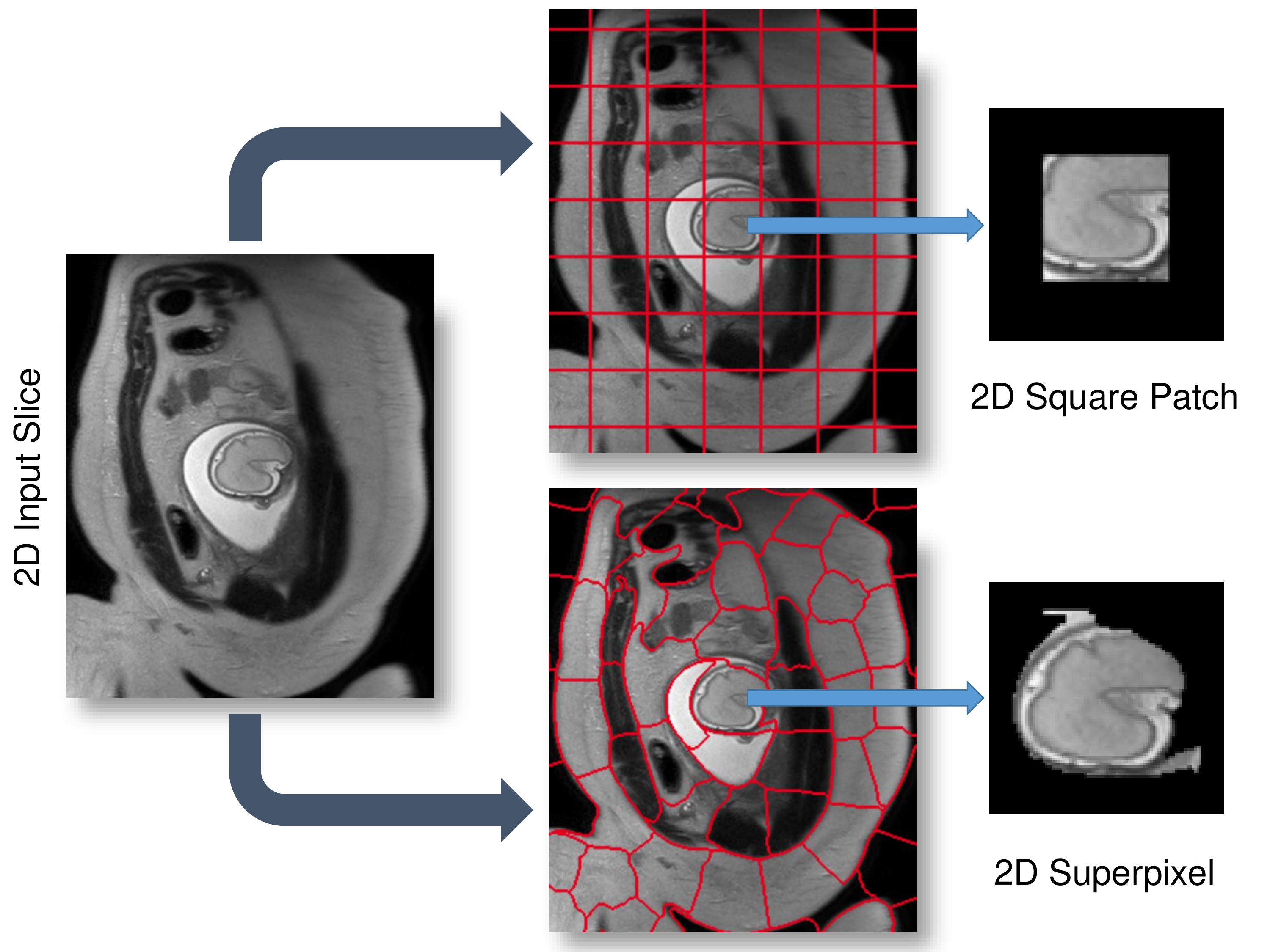}
	\caption[superpixels]{An illustrative figure showing both square patches and superpixels methods for the patch extraction step. A 2D superpixel shows more flexibility than a square patch in extracting rigid regions or similar voxels. In practice, superpixels are dilated with few pixels to include some contextual information in order to increase the accuracy of the patch to volume registration step.}
	\label{fig:spx}
\end{figure}
In the simplest, na\"ive case the shape of $y_s$ is square, and defined via its size $a$ and stride $\omega$. Such definition is generally applicable to any kind of oversampled motion corrupted data. If $a$ and $\omega$ are fixed, no prior knowledge about the data is assumed. However, ideally each $y_s$ corresponds to a meaningful subregion of the volume in which motion can be characterized as rigid. Typically, the square patches are overlapping to provide redundant representations of the same locations. Such approach is computationally expensive with increasing number and patch size $a$ and additional consideration must include the inherent trade-off between $a$ and the assumption of it containing rigid motion.

An alternative to na\"ive shape definitions of $y_s$ is to find correlation between voxel locations and its neighbors, which can be found by unsupervised image segmentation techniques such as \emph{superpixels} (SP)~\cite{achanta2012slic}. These techniques allow for obtaining similar-sized segments from local intensity information (see Fig.~\ref{fig:spx}) instead of employing dense sampling of overlapping patches, enabling the image reconstruction with fewer but more useful data blocks. Further, reducing the total amount of required data blocks for reconstruction lowers the computational overhead, positively impacting the overall run-time. Additionally, larger rigid areas require less computational effort for image registration and super-resolution, and more importantly less dependency on inherent image data parameters (\emph{e.g.}, voxel spacing, organ size, subject size). 

While there are several techniques for generating SP in the literature~\cite{felzenszwalb2004efficient,vedaldi2008quick,achanta2012slic}, a fast and efficient SP approach is desirable for the clinical practice. Simple linear iterative clustering (SLIC)~\cite{achanta2012slic} allows to obtain regular SP based on minimizing the distance $D$ between the centroids of SP with an initial step size $a$. D is defined as:
\begin{equation}
	{D = \sqrt{ {d_c}^2 + {\left({\dfrac{d_s}{a}}\right)}^2 t^2 }}, 
\end{equation}

where $d_c$ and $d_s$ are the intensity and spatial Euclidean distances that are controlled by an adaptive compactness parameter $t$ for each SP. Similarly to na\"ive shape definitions, we can define the initial size of the SP as $a$ and its dilation size $\gamma\%$. 

\textbf{Multi-scale Patches:}
Although larger patch regions are less likely to include rigidly connected regions, they may perform better during 2D-3D registration due to the additional contextual information of each patch. In contrast, smaller patch sizes are more likely to represent rigidly deformed regions, but provide less contextual information, potentially affecting the 2D-3D registration. A good trade-off between the size of the patch region and the likelihood of rigid motion needs to be found. Here, we propose the use of multi-scale patches for reconstruction to exploit the advantages of different patch sizes. We represent input data as stacks of 2D patches:
\begin{equation}
	Y_i = \{y_s | s \in S_i\}, 
\end{equation}
where, instead of using the same $Y$ as a unique input, different scales $Y_i$ are used for each iteration $i$ at $\gamma\%$ of its original size. 
$S_i$ is similar to Eq.~\ref{eq:patch} the set of all locations in all $p$ stacks but with different size for each iteration $i$.
Additionally, to increase contextual information for estimating the transformations, we compute overlapping $y_s$ patches and dilate each superpixel $y_s$ by $\gamma$ pixels using a flat structuring element $b$ with a fixed neighborhood (26 px in our case), hence $\bar{y}_s = y_s \oplus b$. Clearly, the smaller the choice of $\gamma$, the faster is the reconstruction. Ideally $\gamma$ is $>50\%$ of $a$, ensuring that every pixel is covered by multiple samples.

\textbf{Patch to Volume Registration:}
An HR-image $X$ is reconstructed from a number of motion corrupted patches $y_s$ using 2D-3D registration-based super-resolution similar to \cite{murgasova2012reconstruction,kainz2015fast}, where an accurate PSF calculation is used to generate a gradually improving approximation of $X$ and further employed to initialize the 2D-3D registration and computation of robust statistics. In \cite{murgasova2012reconstruction,kainz2015fast}, the PSF equals to a sinc function for the in-plane and the slice profile for the through-plane, measured for the employed MRI sequence (ssFSE), according to \cite{jiang2007mri}. 

\begin{figure}[!htbp]
	\centering 
	\subfloat[]{
		\includegraphics[height=2.55cm]{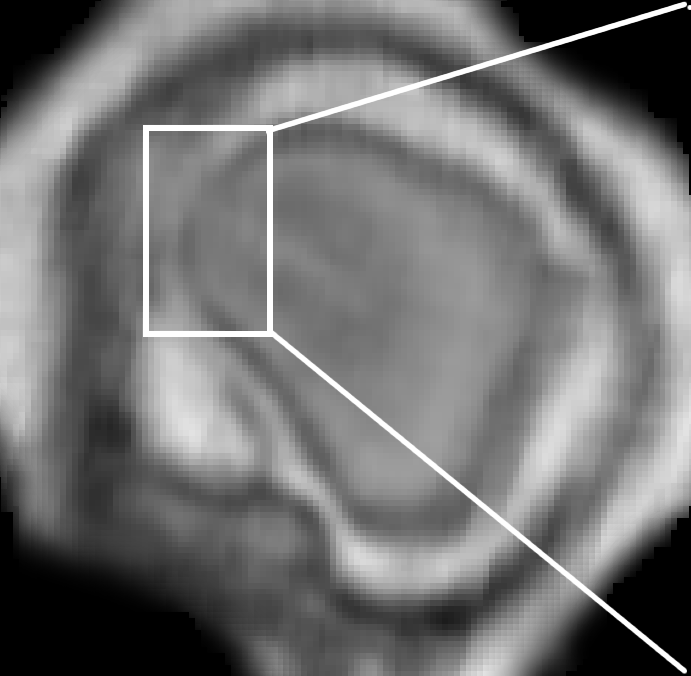}
	}
	\subfloat[]{%
		\includegraphics[height=2.55cm]{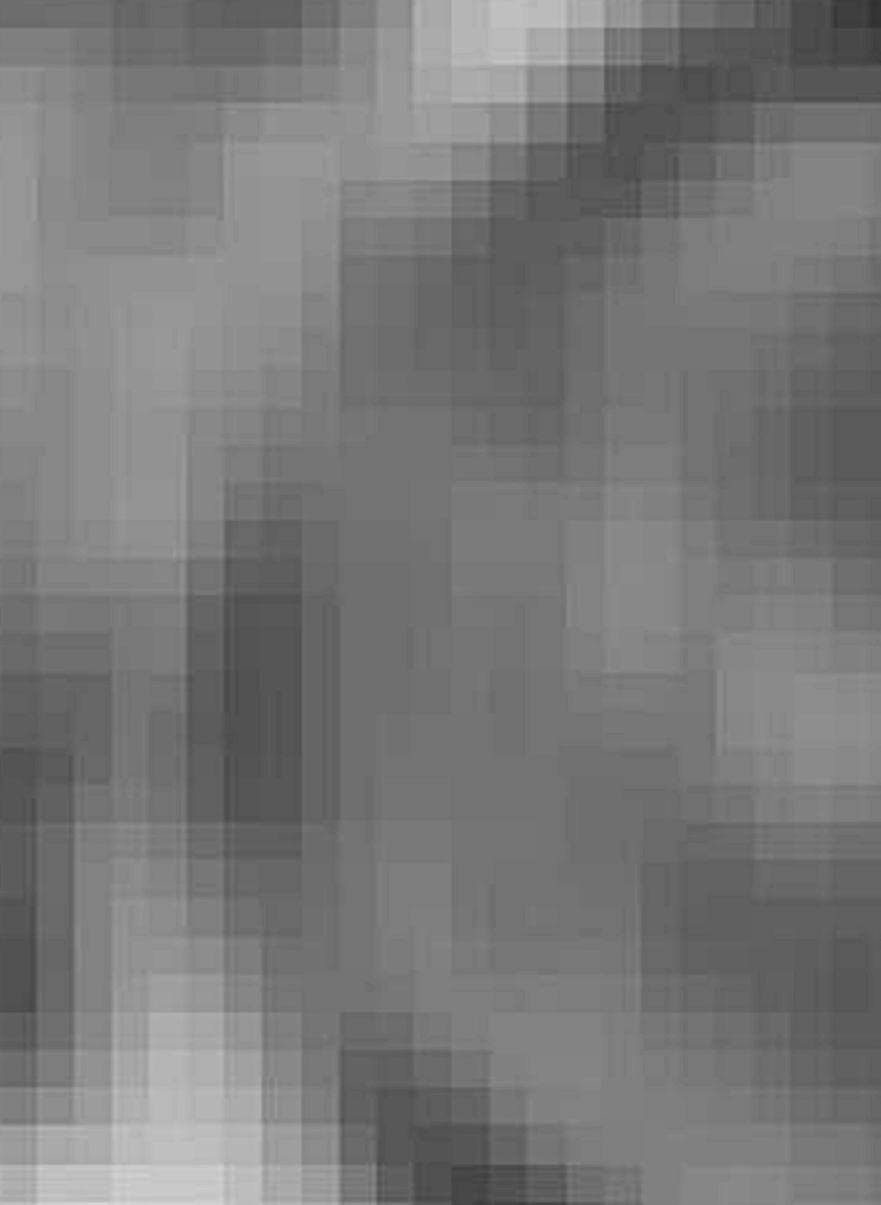}
	}
	\subfloat[]{%
		\includegraphics[height=2.55cm]{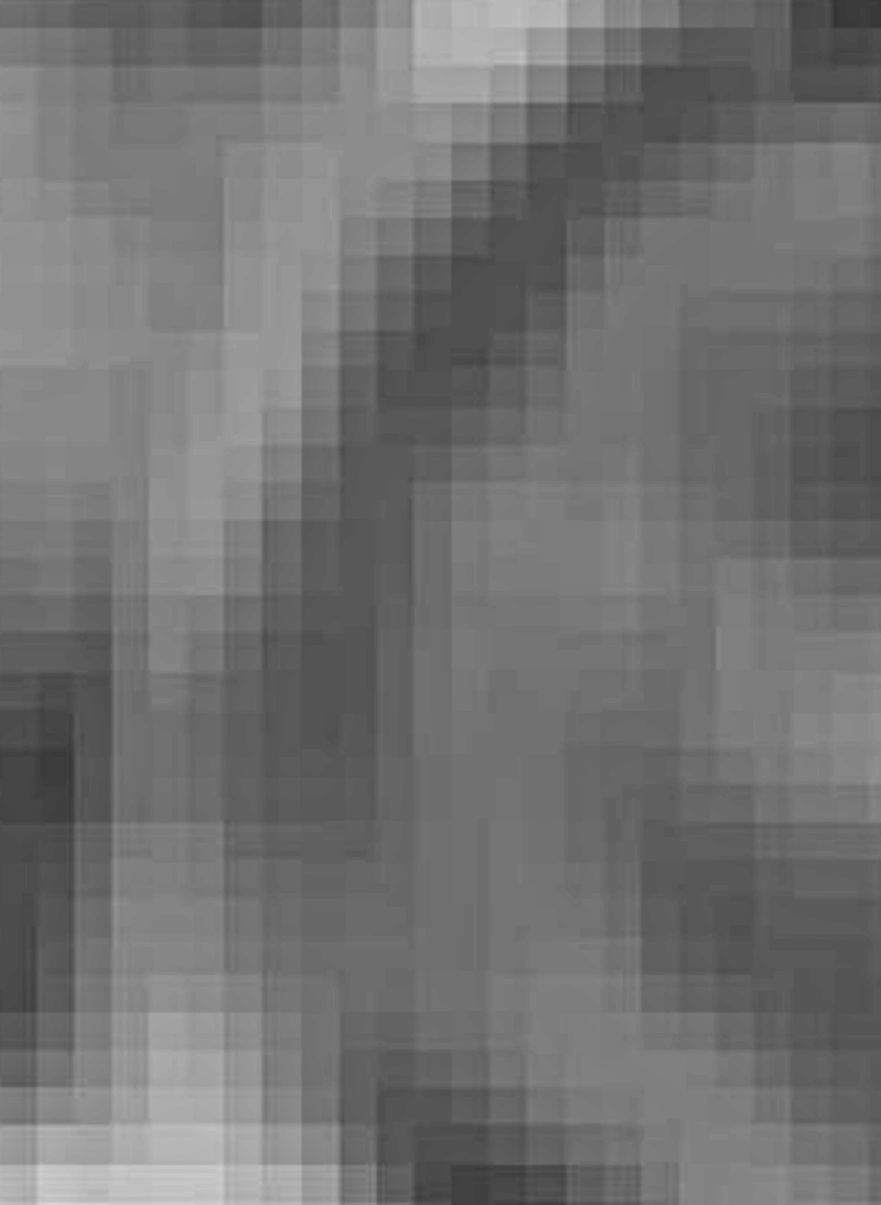}
	}
	\subfloat[]{%
		\includegraphics[height=2.57cm]{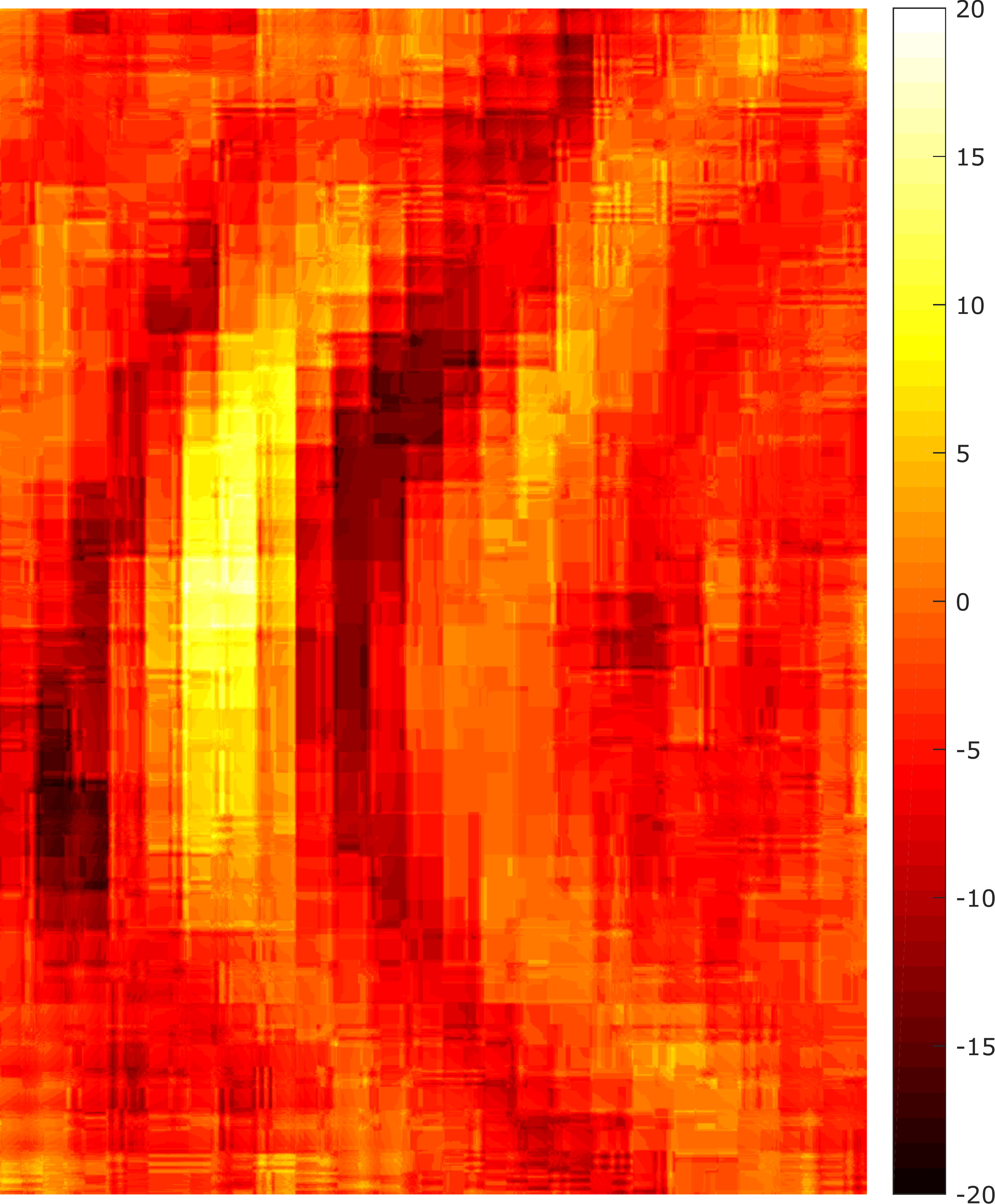}
	}
	\caption{Example for the observed differences in the first iteration of a fetal brain MRI reconstruction (a). (b) shows a magnified region using a sinc function for the PSF similar to~\cite{kainz2015flexible} and (c) shows the result from using a Taylor series approximation of the sinc function as used in this work. Taylor series approximation allows a better approximating of small values close to zero. (d) shows the difference between both images.}
	\label{fig:TaylorPSF}
\end{figure}

We employ an implementation of the PSF function by applying a Taylor series for a better approximating of small values close to $0$. We cut the series after several terms and bound the remainder based on relative error $\epsilon$. The Taylor series approximation of the sinc function is defined as $sinc(R) = 1 - \frac{R^2}{3!} + \frac{R^4}{5!} - \frac{R^6}{7!} + \cdots$. The proposed approximate PSF achieves substantial qualitative improvement in the quality of the reconstructed image compared to the sinc implementation. An example from the first iteration of a fetal brain reconstruction is shown in Fig.~\ref{fig:TaylorPSF}.

During the optimization process, individual 2D patches are continuously rigidly registered to the current 3D reconstruction of $X$ and reintegrated into $X$ using iterative super-resolution with gradient descent optimization. Any similarity metric can be used as a cost function for the registration step such as mutual information~\cite{rousseau2006registration,murgasova2012reconstruction}, cross correlation (CC) \cite{jiang2007mri,kainz2015fast}, or mean square intensity differences~\cite{kim2010intersection,gholipour2010robust}. Choosing the best similarity metric for reconstruction depends on the input data. 
CC has been found to be effective for input data with similar intensity distribution~\cite{penney1998comparison}. In our experiments, we employ CC as the similarity metric for 2D-3D registration, after rescaling the intensities between the input stacks.

\textbf{EM-Based Outlier Removal:} 
Correctly registered patches $\bar{y}_s$ should provide a higher contribution to the final reconstruction, presenting with a low error $e$ when compared to the original image data. \cite{gholipour2010robust} initially introduced an approach to account for outliers during super-resolution based on Huber function statistics. Similar to \cite{murgasova2012reconstruction}, we employ the expectation maximization algorithm for outlier removal by classifying $\bar{y}_s$ and the included pixels into an inlier and outlier class. A zero-mean Gaussian distribution $G_\sigma(e)$ with variance $\sigma^2$ is used for the inliers and a uniform distribution with constant density 

\begin{equation}
	m = \frac{1}{max(e)-min(e)}
\end{equation}
for the outliers. This makes use of available, highly redundant information (\emph{i.e.}, overlapping $\bar{y}_s$), to find partly matching patches and to depreciate or fully reject erroneous voxels. We aim to maximize the log-likelihood for each patch 
\begin{equation}
	{y}_s | log P(Y,\Phi) = \sum log P(e|\sigma,c)
\end{equation}
to be part of a region of rigid motion. $\Phi$ is the current estimate of the reconstructed volume $X$, the variance $\sigma^2$ of the errors $e$, and the proportion of correctly matched voxels $c$. The posterior probability for a pixel $\in \bar{y}_s$ being identified as inlier is 

\begin{equation}
	p = \frac{G_\sigma(e)c}{G_\sigma(e)c+m(1-c)}.
\end{equation}
We perform the updates of $c$ and $\sigma^2$ according to \cite{murgasova2012reconstruction}:

\begin{equation}
	\bar{p} = \sqrt{(\sum_{\bar{y}_s} p^2)/N},
\end{equation}
where $N$ is the number of pixels in $\bar{y}_s$. We further define an inlier and outlier probability for each $\bar{y}_s$ and exclude it from processing if classified as an outlier (\emph{e.g.}, if it contains structures moving in opposite directions during scanning, such as the fetal head and thorax). Only if information in $\bar{y}_s$ is consistent with the originally acquired data, the registered patch will remain contributing to the SR reconstruction of $X$.

\textbf{Identification of Rigid Regions and SVR Refinement:} 
The \emph{rigidity} of regions is measured by keeping track of the probability $p$ of each pixel of every $\bar{y}_s$. This allows to identify locations best fitting the rigid 2D-3D registration constraints. Integrating $p$ and $\bar{p}$ into a 3D volume using the same PSF as for the reconstruction identifies candidate regions, solely containing rigid motion components~\cite{kainz2015flexible}. This can further be used to initialize the rigid SVR reconstruction or to visualize the data uncertainty during reconstruction.

\section{Implementation}

\textbf{Parallelization:}
\label{sect:Parallelization}
The high data redundancy required for the proposed approach makes conventional single threaded implementation practically not feasible. Computational complexity of PVR is exponentially higher than SVR, depending on the employed patch overlap. For optimal performance we implemented our approach via General-Purpose Programming on Graphics Processing Units (GPGPU) using the Compute Unified Device Architecture (CUDA, NVIDIA, Santa Clara, CA) language \cite{Nickolls2008,Sanders2010}. CUDA is a highly evolved single instruction multiple data (SIMD) programming language, which allows a large part of the proposed framework to be mapped onto GPU hardware. Currently, CUDA is the only high-level general purpose GPU language that provides, for example, bi-directional texture access via surfaces in a kernel, which is essential for the efficient implementation of certain parts our framework. In this section we discuss the key implementation details.

\begin{figure*}[htbp]
	\centering
	\includegraphics[width=\textwidth]{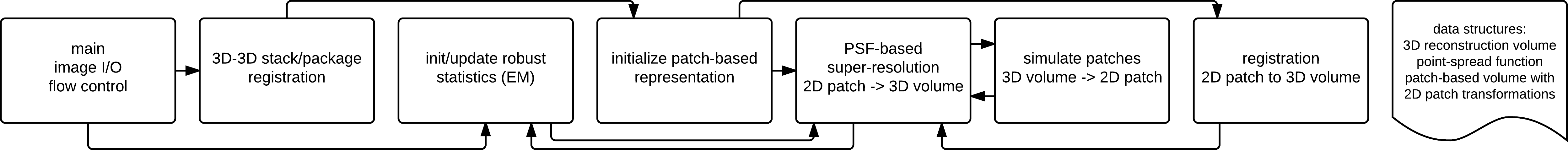}
	\caption[SWOverview]{The software modules defined for the implementation of the proposed approach. For implementation details, please refer to the provided source code.}
	\label{fig:SWoverview}
\end{figure*}

We use a modular design to allow experimentation with the separate components of the algorithm. An overview of this design is shown in Fig. \ref{fig:SWoverview}. The modules are encapsulated in a CUDA library, which can be used independently from the instantiating framework. We employ the successor of IRTK\footnote{\url{https://github.com/BioMedIA/MIRTK}} for interfacing with medical image data. 

PVR is parallelized on three levels:
\begin{enumerate} 
	\item[I.]\emph{Patch-level:} Individual patches are mapped to blocks of a CUDA computing grid and the contained voxels are mapped to individual threads. Depending on the used GPU hardware, patch processing can be also mapped directly to the computing grid, such that each thread works on a complete patch (limited by the employed patch size). The resulting thread divergence provides opportunities for advanced GPU scheduling strategies~\cite{Steinberger2012} and for a direct translation of optimization strategies for image registration, for example patch-wise gradient descent.
	\item[II.]\emph{Voxel-level:} For the parallelization of PSF-based super-resolution and robust statistics we follow a similar three-folded procedure definition as used in~\cite{kainz2015fast}. The voxels within each patch are processed using kernel level parallelization and parallel pixel-volume, volume-pixel, and volume-volume procedures are applied.
	\item[III.]\emph{Patch-batch:} PVR scales to multiple GPUs through distributing independent subsets of patches over the desired number of devices. Synchronisation is done through averaging of the resulting sub reconstruction volumes on the master GPU.
	 Initial 3D-3D registration is performed on a single master GPU, which allows optimal coalesced memory access.
\end{enumerate}

\textbf{Availability of Source Code: }
We provide the source code of a c++/CUDA implementation of the proposed method, including parallelization strategies (see Sec. \ref{sect:Parallelization}), in a publicly available software repository\footnote{\url{https://github.com/bkainz/fetalReconstruction}}. 
The source code for the implementation of PVR is licensed under MIT license. 

\section{Evaluation \& Experiments}
\label{sec:eval_exp}

\textbf{Evaluation of Adult Brain MRI Reconstruction:}
\label{subsec:adult_exp}
We evaluate the performance and limitations of PVR in terms of accuracy and robustness  with 
 synthetic non-rigid deformations of \emph{adult} brain data. Similar to~\cite{gholipour2010robust}, an isotropic $1$ $mm^3$ T2-weighted adult brain phantom with no noise obtained from the \emph{Brainweb} database~\cite{Cocosco1997} is used for this experiment. Synthetic non-rigid motion artifacts are generated by skewing (shearing) the original image using: 
\[
	T_{\theta_{xyz}} = 	\begin{bmatrix}
						1 		   &S_{xy} 	&S_{xz} 	&0 	\\
						S_{yx} 	 &1			  &S_{yz}		&0 	\\
						S_{zx}	 &S_{zy}	&1			  &0	\\
						0		     &0			  &0			  &1
						\end{bmatrix},
\]
\noindent where we use one combined skewing value in the $xyz$-direction defined by $S_{xyz}=tan(\pm\theta_{xyz}^{\circ})$.
After that a motion-corrupted 3D stack is constructed by sampling 2D images from both skewed and motion-free stacks in an interleaved manner similar to fetal MRI acquisition~\cite{glenn2006magnetic}. Three stacks are used for the reconstructions where each stack is sampled with a voxel size of $1.25$x$1.25$x$2.5$ $mm^3$. We use standard axial, sagittal, and coronal orientations as shown in Fig.~\ref{fig:sim_motion}. An HR image with isotropic voxel size $1.25$ $mm^3$ is reconstructed using SVR \cite{kainz2015fast}, square patch- and superpixel-based PVR. 

\textbf{Evaluation of Fetal Organ MRI Reconstructions:}
\label{subsec:fetal_exp}
Evaluating the quality of reconstructed fetal MRI is challenging due to the absence of motion-free ground truth data. For this purpose, we introduce a novel approach for this evaluation problem based on the originally acquired slice images. Assuming that 2D in-plane patches extracted from the original stacks contain no motion artifacts, we use them as gold standard and compare them with corresponding simulated patches from the reconstructed volume.
Evaluation metrics (see Sec. \ref{subsec:metrics}) are computed between the reconstructed input stacks and the final motion corrected image and averaged over the whole volume. 
The fetal brain is typically used to assess the quality of reconstruction as it moves rigidly, fulfilling the rigid motion assumption for SVR-based methods in the 2D-3D registration step. However, soft tissue organs such as the placenta deform non-rigidly. For this reason, we additionally chose to reconstruct the placenta and the whole uterus as challenging test cases for PVR and SVR.

\subsection{Evaluation metrics}
\label{subsec:metrics}
We employ the following metrics for measuring the quality of the reconstructed image:	
\noindent\uline{Cross-correlation (CC)} to measure the similarity between the intensities of input $I(i,j)$ and reconstructed image $\tilde{I}(i,j)$ at the location $(i,j)$, which is defined as:
		\begin{equation}
			CC = \frac{1}{N \times M} \sum_{i=1}^{N} \sum_{j=1}^{M} \frac{ (I(i,j)-I_{\mu}) (\tilde{I}(i,j)-\tilde{I}_{\mu}) } { \sigma_{I} \sigma_{\tilde{I}} }
			\label{equ:cross_corr}
		\end{equation}
		\noindent where N and M are the dimensions of a 2D slice. 
        
\noindent The \uline{peak signal-to-noise ratio (PSNR)} is used to measure the error introduced by motion and is based on the mean squared error (MSE) between the original 2D in-plane patch and the reconstructed image. PSNR is defined as:
		\begin{equation}
			PSNR = 10 \log \frac{I_{max}^{2}}{MSE}
			\label{eqn:PSNR}
		\end{equation}
		\noindent where $I_{max}$ is the maximum intensity in the original image. An improved reconstruction quality usually results in higher PSNR. However, PSNR does not reflect well subjective human perception of image quality as it is mainly based on estimating absolute errors between individual pixels. 

\noindent The \uline{structural similarity index (SSIM)} accounts for image degradation as perceived changes in structural information~\cite{wang2004image}. It measures the structural similarity by comparing normalized local patterns of pixel intensities, which is similar to the human visual system's abilities to extract information based on structure. The SSIM is defined as:
\begin{equation}
	SSIM = \frac{(2\mu_{I}\mu_{\tilde{I}} + c_{1}) (2\sigma_{I\tilde{I}} + c_{2} )} {(\mu_{I}^2 \mu_{\tilde{I}}^2 + c_{1}) (\sigma_{I}^2 + \sigma_{\tilde{I}}^2 + c_{2}) }
	\label{eqn:SSIM}
\end{equation}
\noindent where $\mu_{I}$, $\mu_{\tilde{I}}$, $\sigma_{I}^2$ and $\sigma_{\tilde{I}}^2$ are the average and variance of the intensities of the original 2D in-plane slice and the reconstructed slice respectively. $\sigma_{I\tilde{I}}$ is the covariance of $I$ and $\tilde{I}$. $c_{1}$ and $c_{2}$ are defined as $(k_{1}L)^2$ and $(k_{2}L)^2$ in order to balance the division with weak denominator, where $L$ is the dynamic range of the intensities in image $I$ and $k_{1}$. $k_{2}$ equal to $0.01$ and $0.03$ similar to~\cite{wang2004image}.

\noindent \uline{Structural dissimilarity (DSSIM)} heat maps are calculated in order to visualize the dissimilarities between original and reconstructed images. DSSIM is calculated as a distance metric derived from SSIM:
		\begin{equation}
		DSSIM = \frac{(1-SSIM)}{2}
		\label{eqn:DSSIM}
		\end{equation}

\section{Results}

\begin{figure}[!htbp]
	\centering 
	\subfloat{
		\raisebox{0.3 in}{\rotatebox[origin=t]{90}{(a) Input}} 
	}
	\hfill
	\subfloat{%
		\includegraphics[width=3.1in,height=0.75in, trim=0 0 0 0,clip]{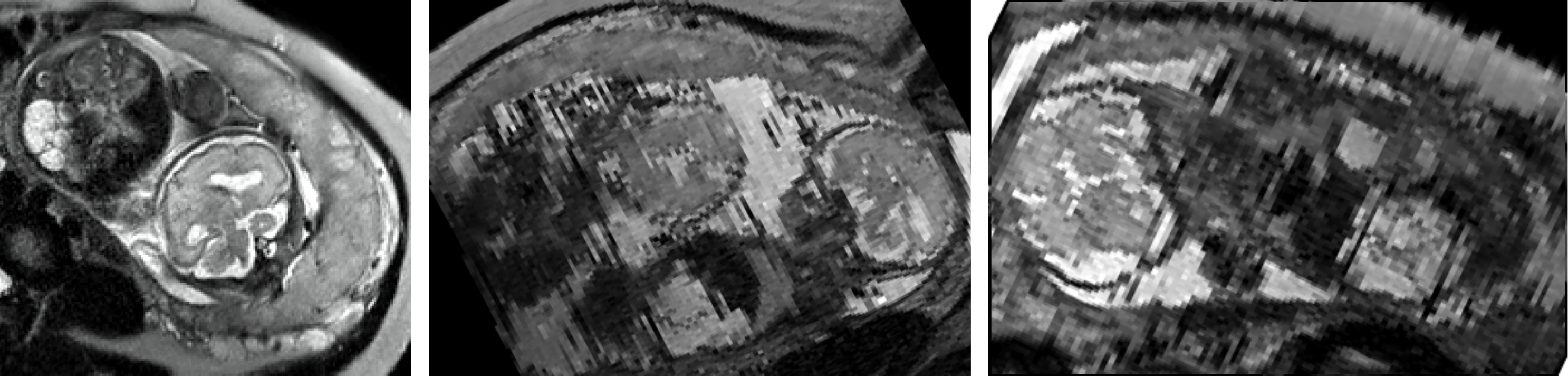}
	}
	\hfill\null
	\\
	\hfill
	\subfloat{%
		\raisebox{0.3 in}{\rotatebox[origin=t]{90}{(b) PVR}}
	}
	\hfill
	\subfloat{%
		\includegraphics[width=3.1in,height=0.75in, trim=0 0 0 0,clip]{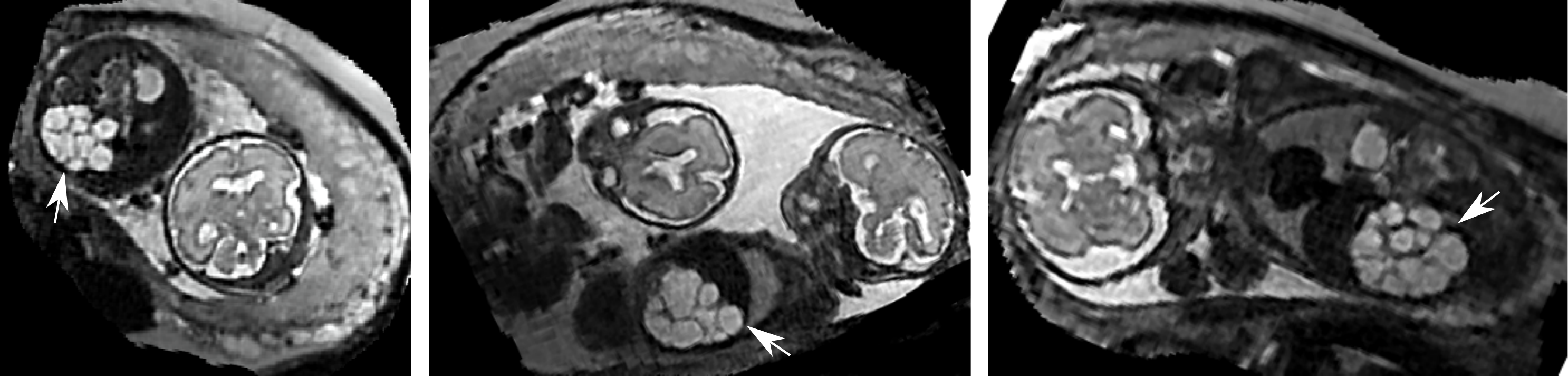}
	}
	\hfill\null
	\caption{Three viewing planes through the originally scanned (a) and the reconstruction (b) of a motion corrupted scan from moving twins with a gestational age of 28 weeks using multi-scale superpixels. For this dataset we used a mask of the uterus to save unnecessary computation time in areas containing maternal tissue. The white arrow points at a unilateral multicystic kidney of one of the twins.}
	\label{fig:twins}
\end{figure}

\begin{figure}[htbp]
\centering
	\null\hfill
	{%
		\includegraphics[width=1.0\columnwidth, trim=0 0 0 0,clip]{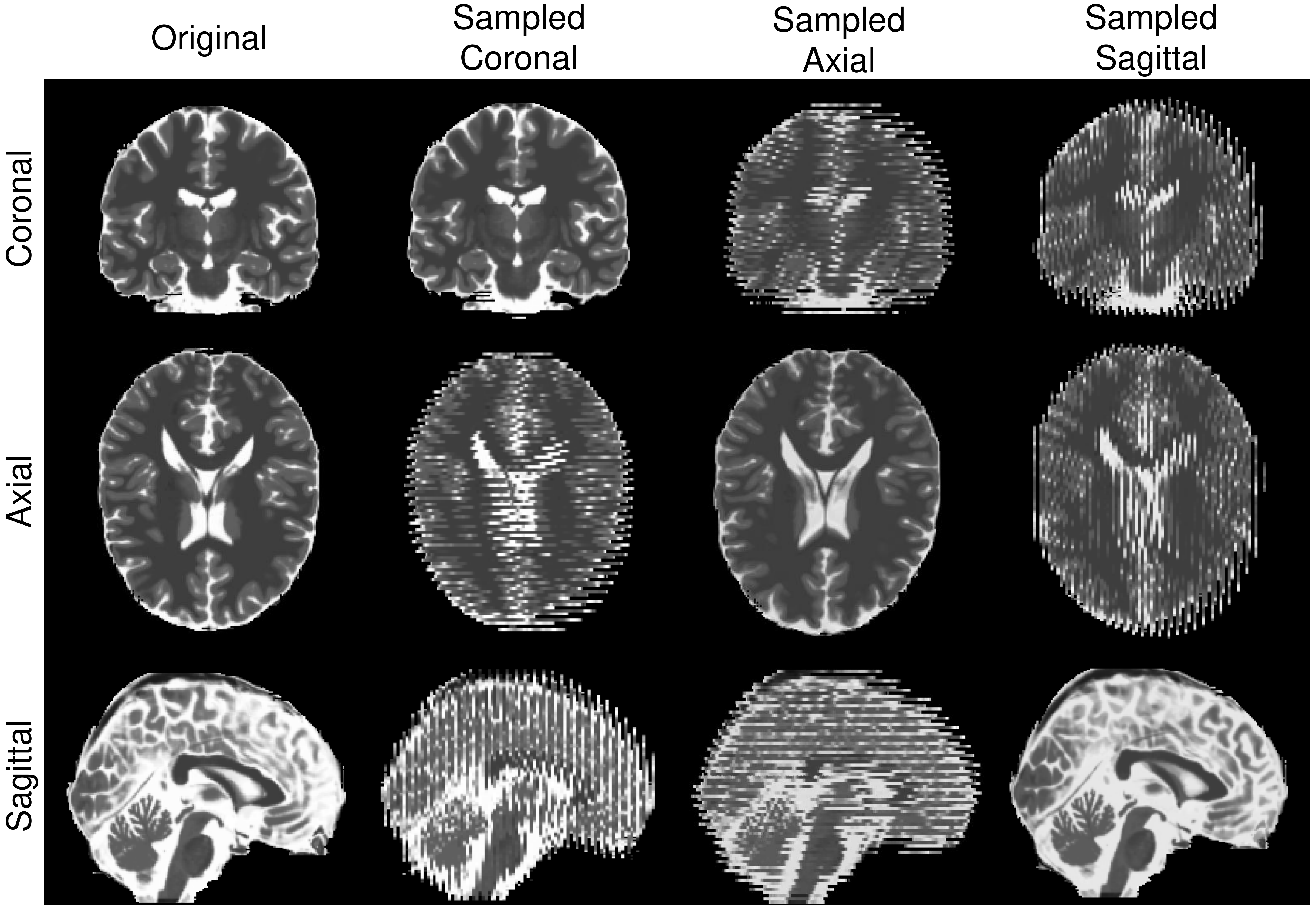}
	}
	\hfill\null
	\caption{Synthetic motion artifacts caused by skewing on the Brainweb Adult MRI Phantom \cite{Cocosco1997}. Rows: MRI in standard orientations: coronal, axial, and sagittal. Columns: original, coronally, axially and sagitally sampled.}
\label{fig:sim_motion}
\end{figure}

\begin{figure}[htbp]
\centering
	{%
		\includegraphics[width=1.0\columnwidth, trim=10 35 10 10,clip]{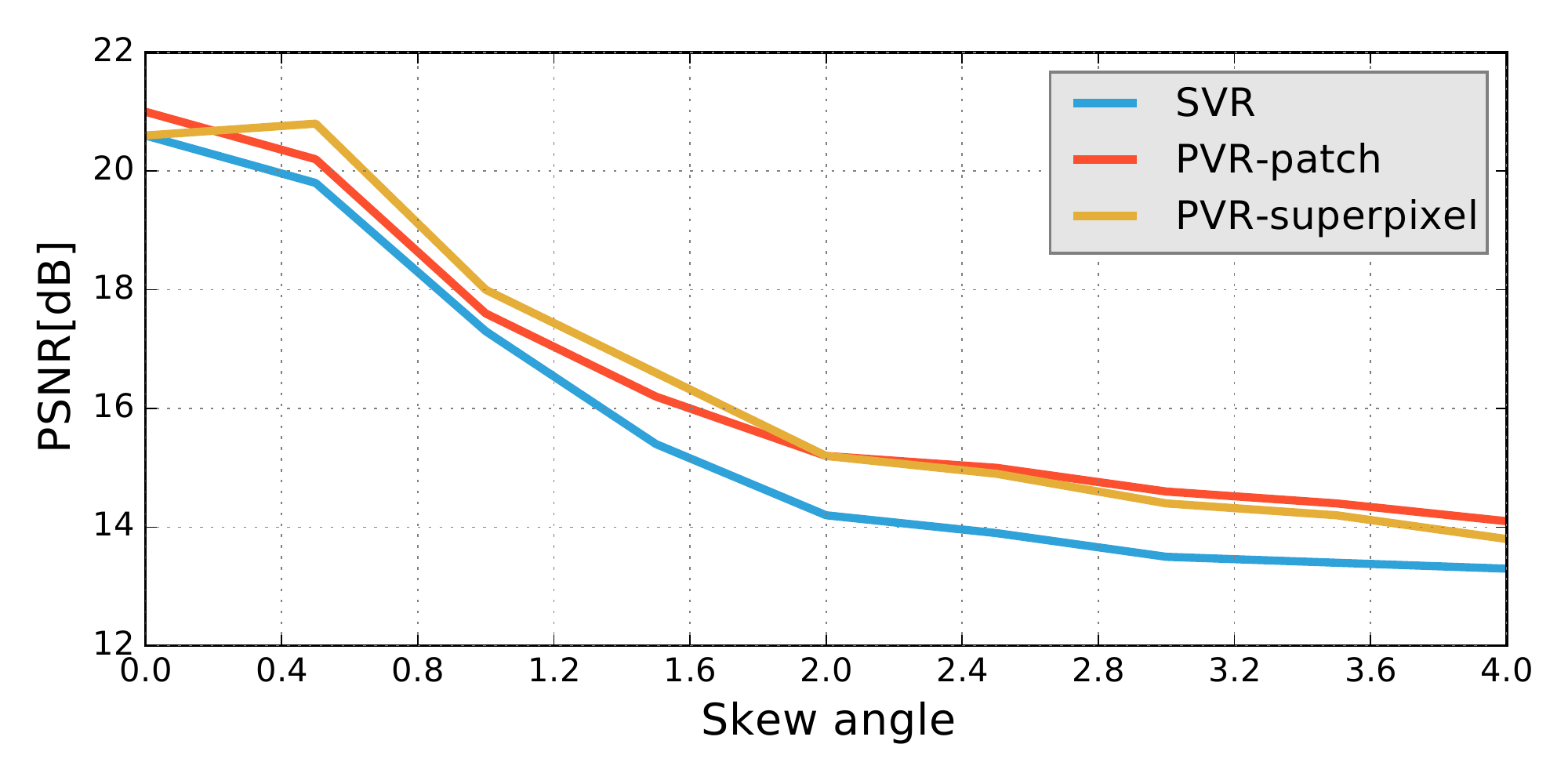}
	}
	\\
	{%
		\includegraphics[width=1.0\columnwidth, trim=10 35 10 10,clip]{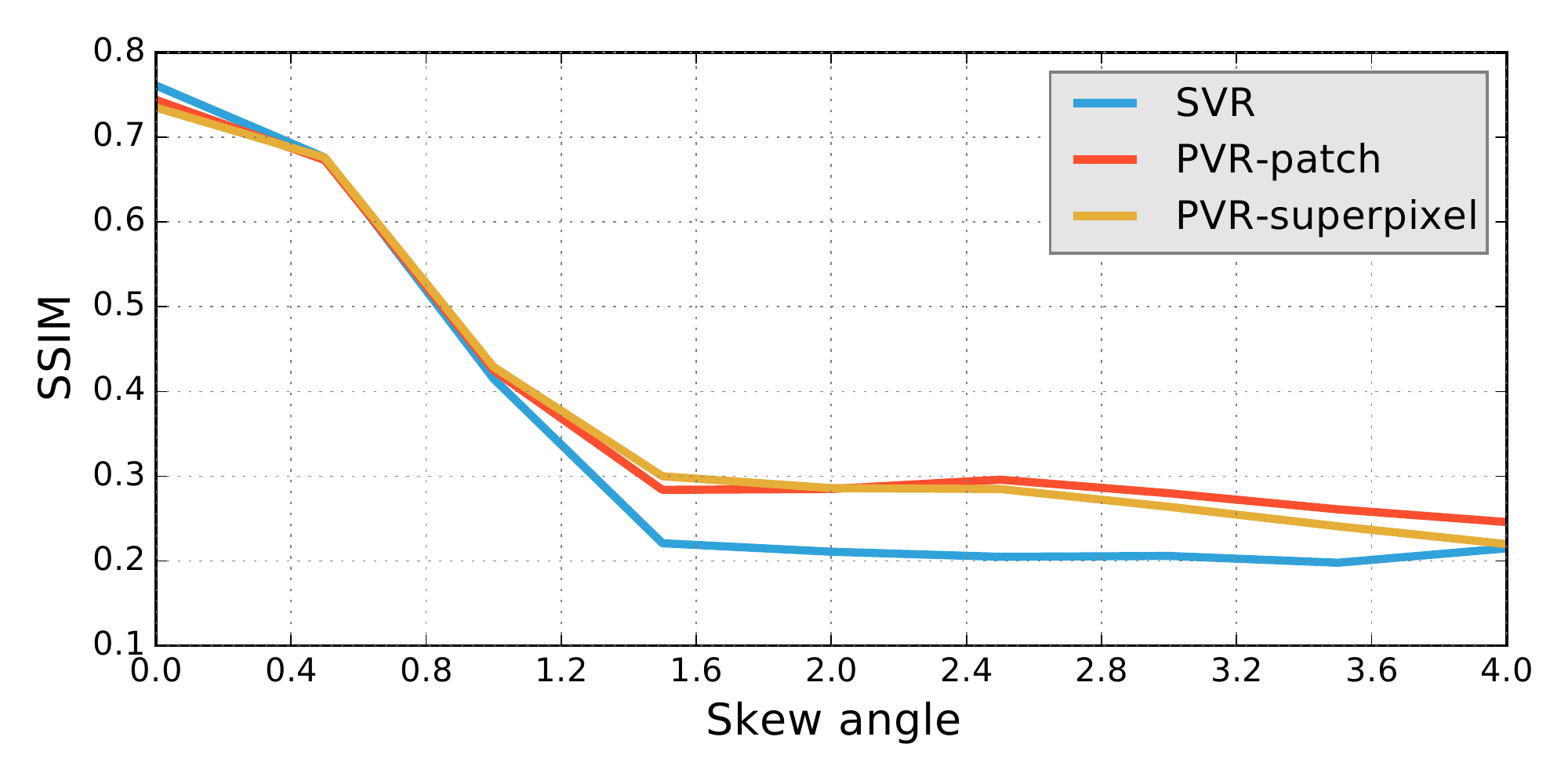}
	}
	\\
	{%
		\includegraphics[width=1.0\columnwidth, trim=10 10 10 10,clip]{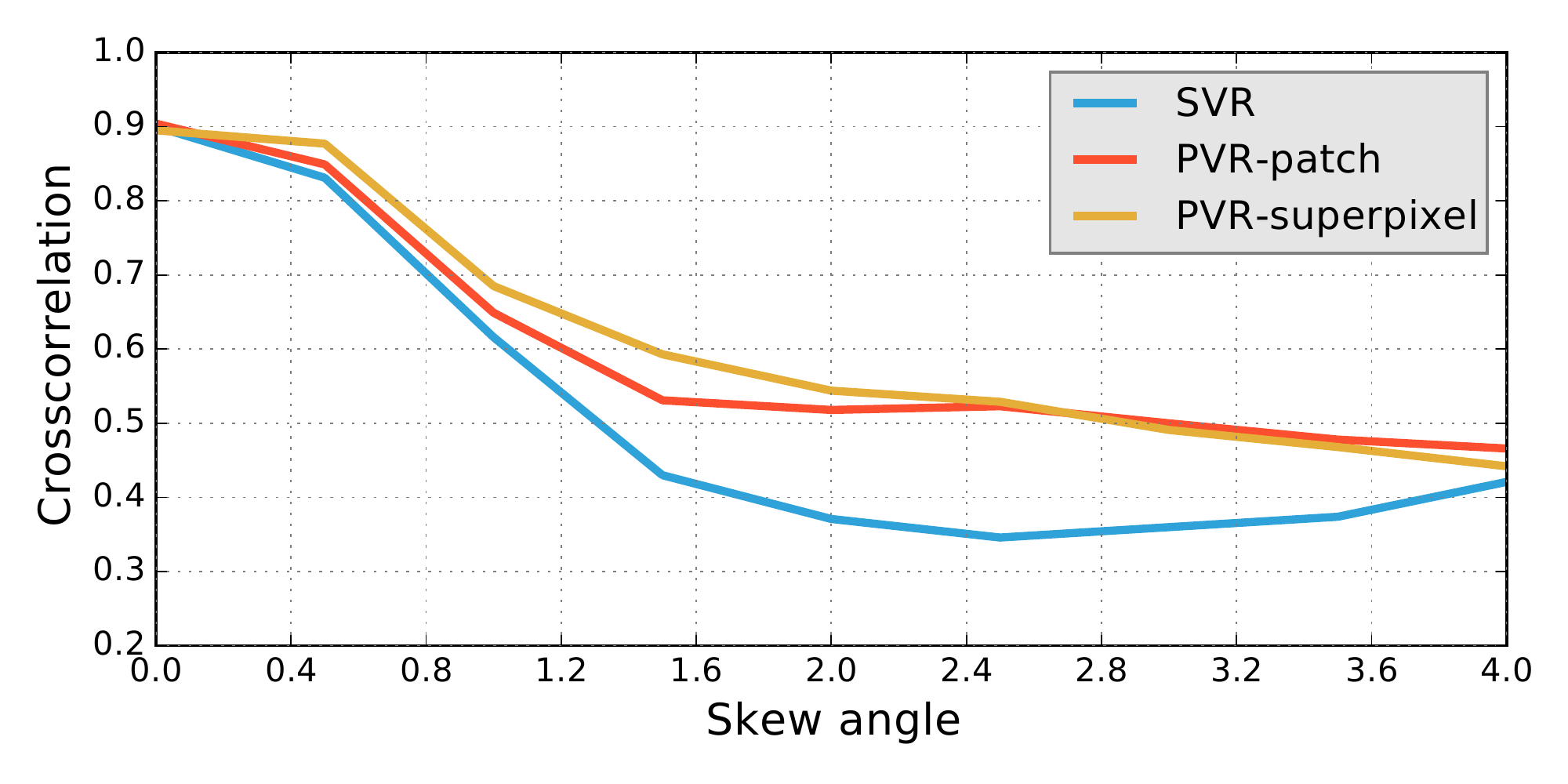}
	}
	\caption{Comparative reconstruction performance of SVR and PVR methods on synthetically corrupted Brainweb \cite{Cocosco1997} data. Top to bottom: PSNR, SSIM and CC over skew angle for SVR (blue), superpixel-based PVR ($a=16, \gamma=60\%$, yellow) and PVR using square patches ($a=32, \omega=16$, red).}
\label{fig:svr_pvr}
\end{figure}

\textbf{Reconstruction of Adult Brain MRI:}
\label{subsec:adult_res}
Experiments on adult brain MR data using the Brainweb database \cite{Cocosco1997} included introducing synthetic non-rigid motion artifacts as described in Sec. \ref{subsec:adult_exp}. Example slices of standard planes of original and corrupted data are depicted in Fig. \ref{fig:sim_motion}. Comparative experimental results of SVR and PVR reconstruction methods are shown in Fig. \ref{fig:svr_pvr} for PSNR, SSIM and CC. 
For all metrics, PVR shows an improved performance over SVR, particularly in presence of deformations with higher skewing angles. Further, we observe that superpixel-based PVR achieves similar performance as PVR using arbitrary square patches, while requiring a lower amount of input patches. 

\begin{figure}[!htbp]
\centering
	\null\hfill
	\subfloat{
		\raisebox{0.35in}{\rotatebox[origin=t]{90}{(a) Input}} 
	}
	\hfill
	\subfloat{%
		\includegraphics[width=0.27\columnwidth,angle=90, trim=0 0 0 0,clip]{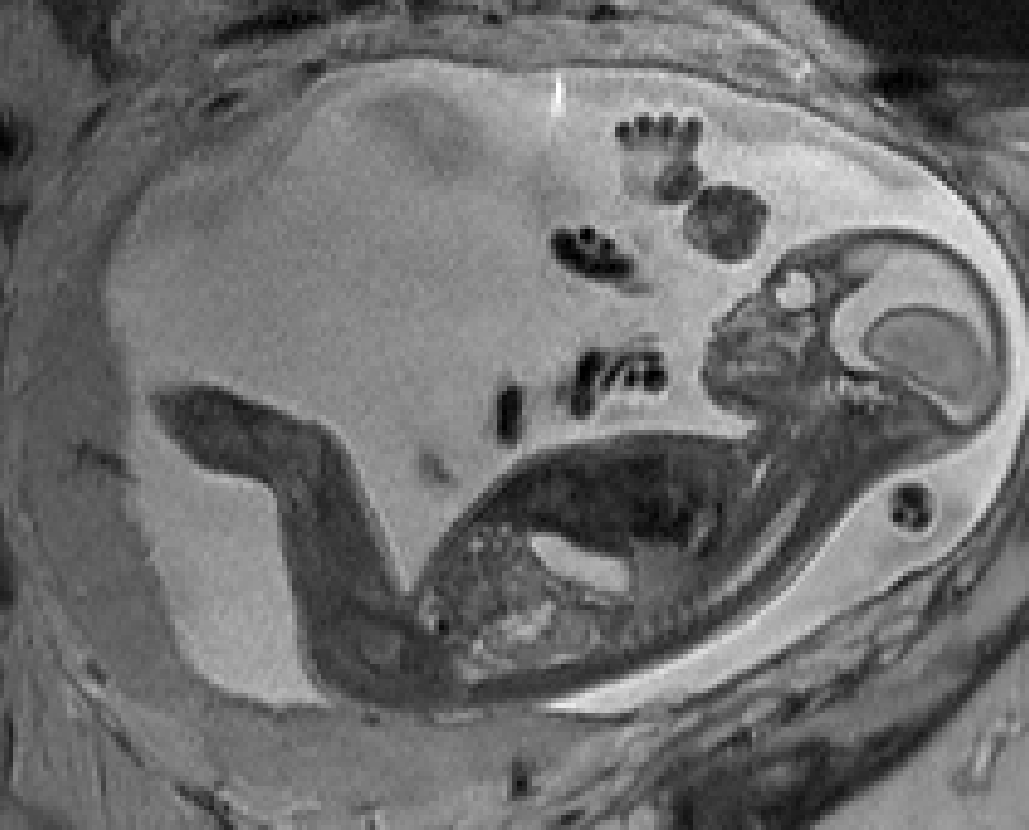}
	}
	\hfill
	\subfloat{%
		\includegraphics[width=0.27\columnwidth,angle=90, trim=0 0 0 0,clip]{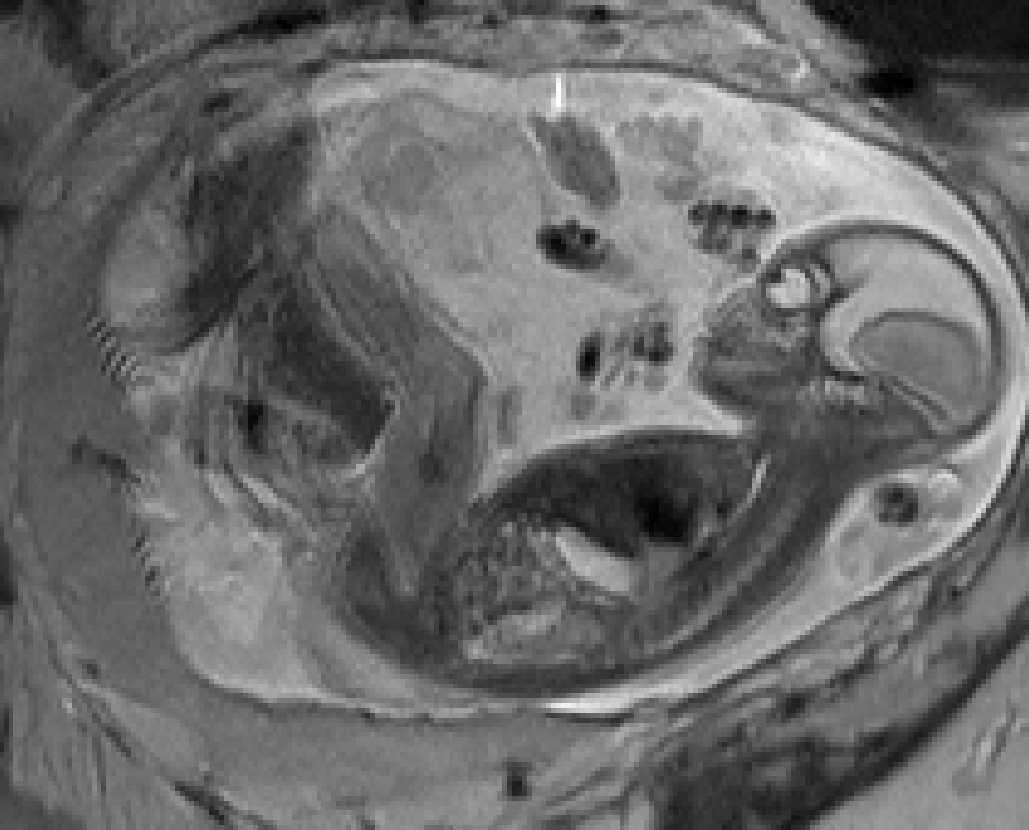}
	}
	\hfill
	\subfloat{%
		\includegraphics[width=0.27\columnwidth,angle=90, trim=0 0 0 0,clip]{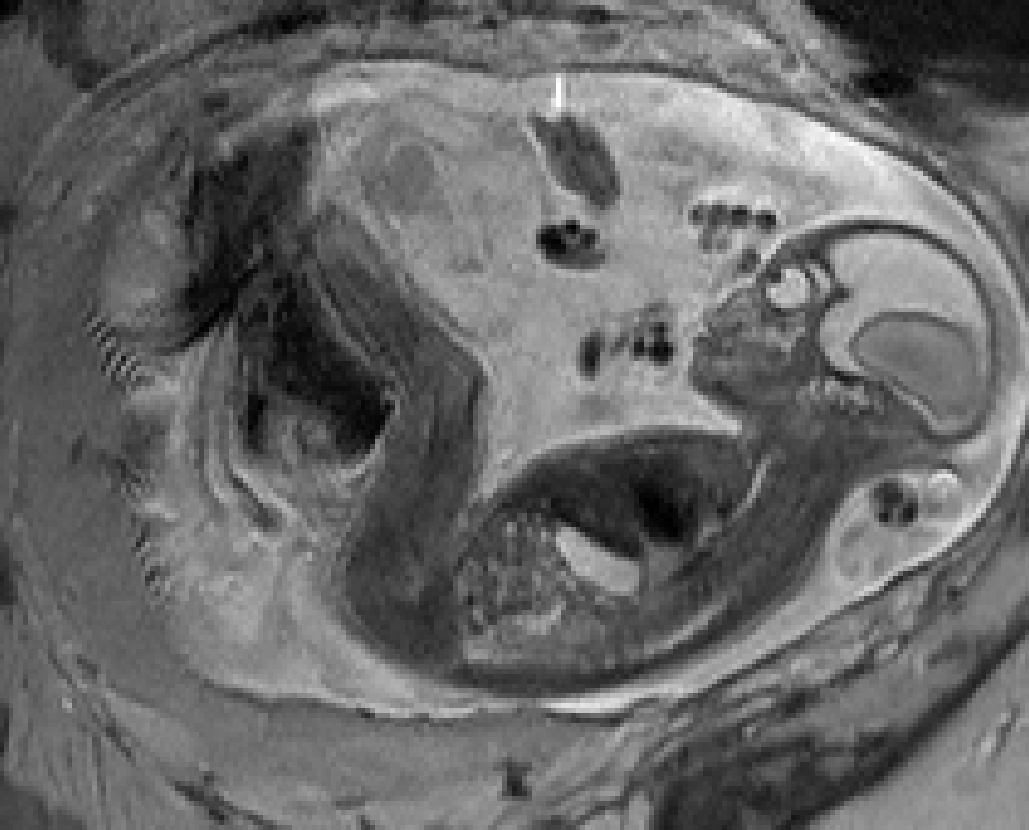}
	}
	\hfill
	\subfloat{%
		\includegraphics[width=0.27\columnwidth,angle=90, trim=0 0 0 0,clip]{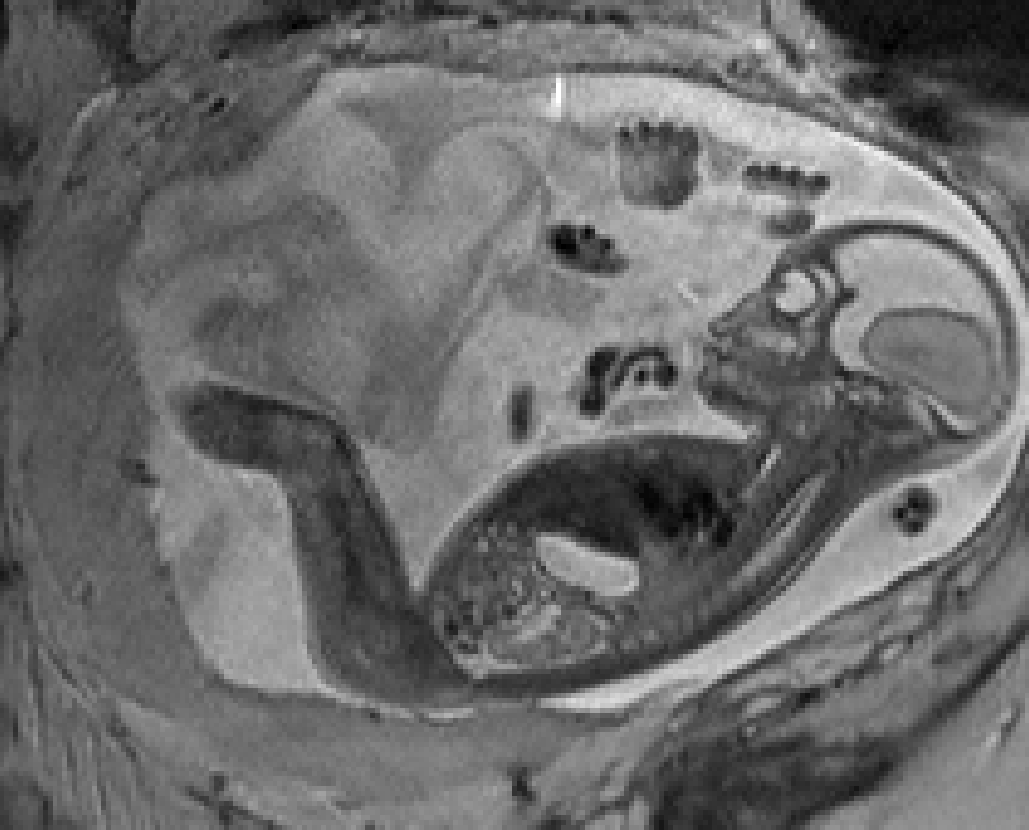}
	}
	\hfill
	\\
	\hfill
	\subfloat{
		\raisebox{0.35in}{\rotatebox[origin=t]{90}{(b) SVR}} 
	}
	\hfill
	\subfloat{%
		\includegraphics[height=0.27\columnwidth,width=0.22\columnwidth,angle=0, trim=80 140 20 80,clip]{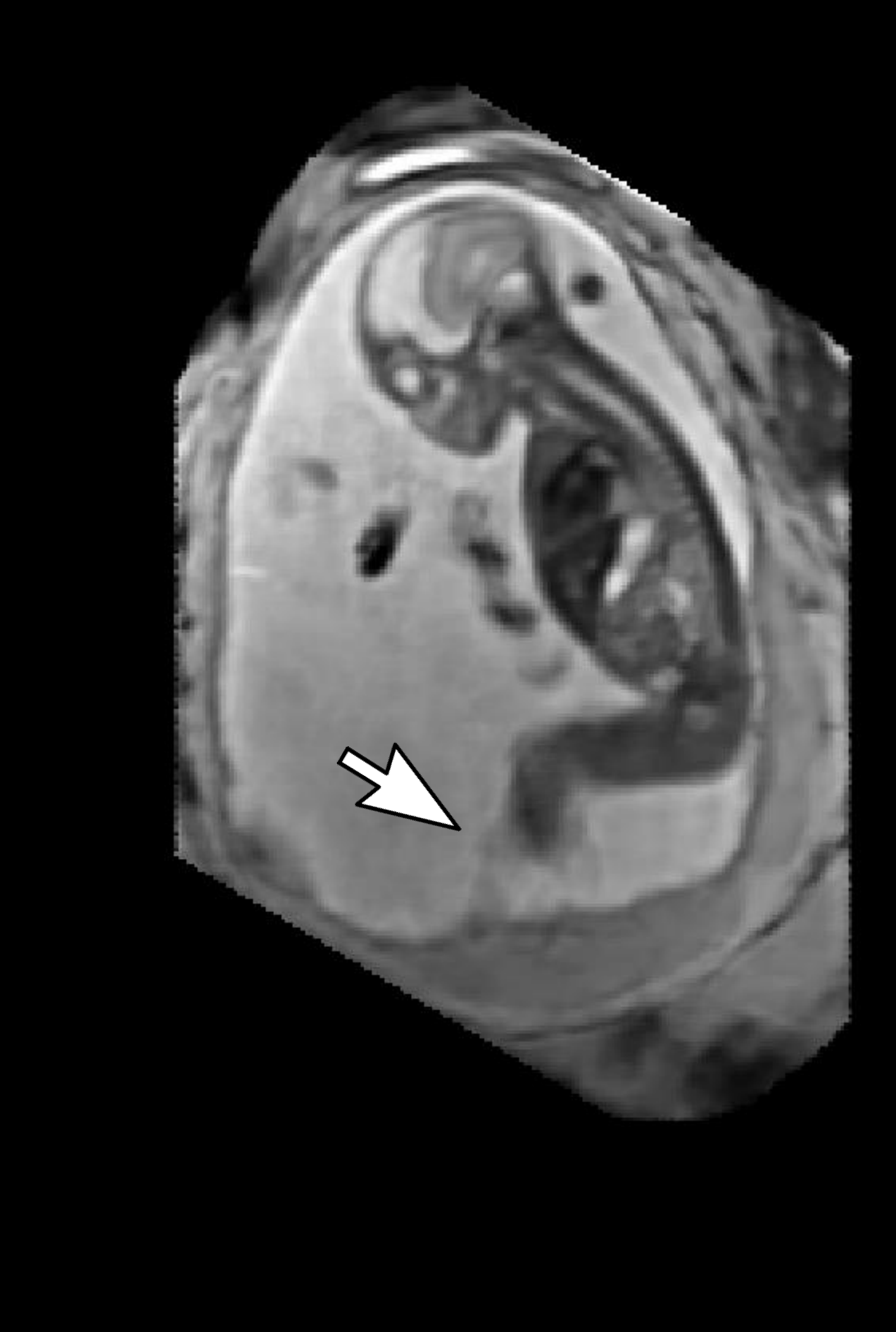}
	}
	\hfill
	\subfloat{%
		\includegraphics[height=0.27\columnwidth,width=0.22\columnwidth,angle=0, trim=80 140 20 80,clip]{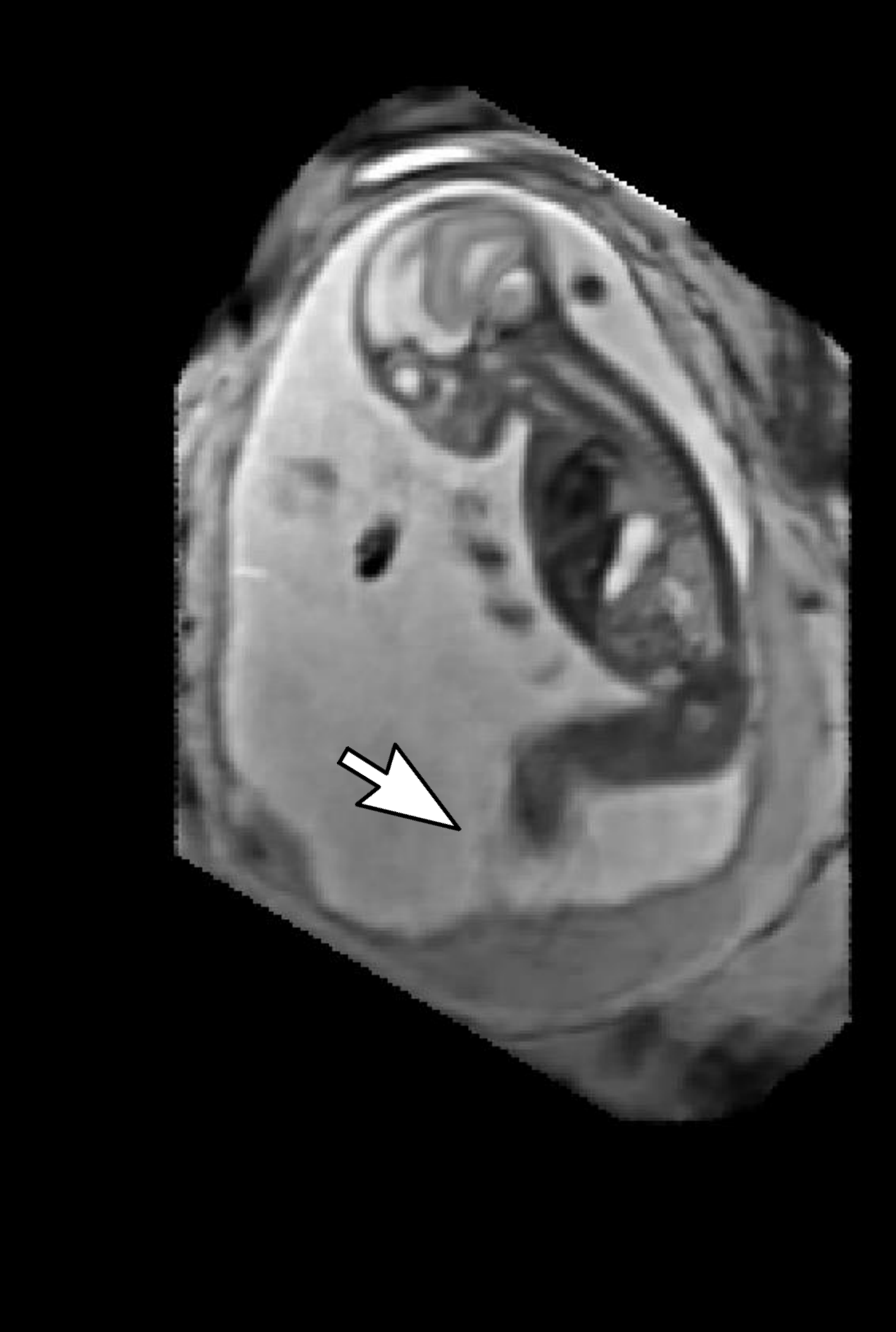}
	}
	\hfill
	\subfloat{%
		\includegraphics[height=0.27\columnwidth,width=0.22\columnwidth,angle=0, trim=80 140 20 80,clip]{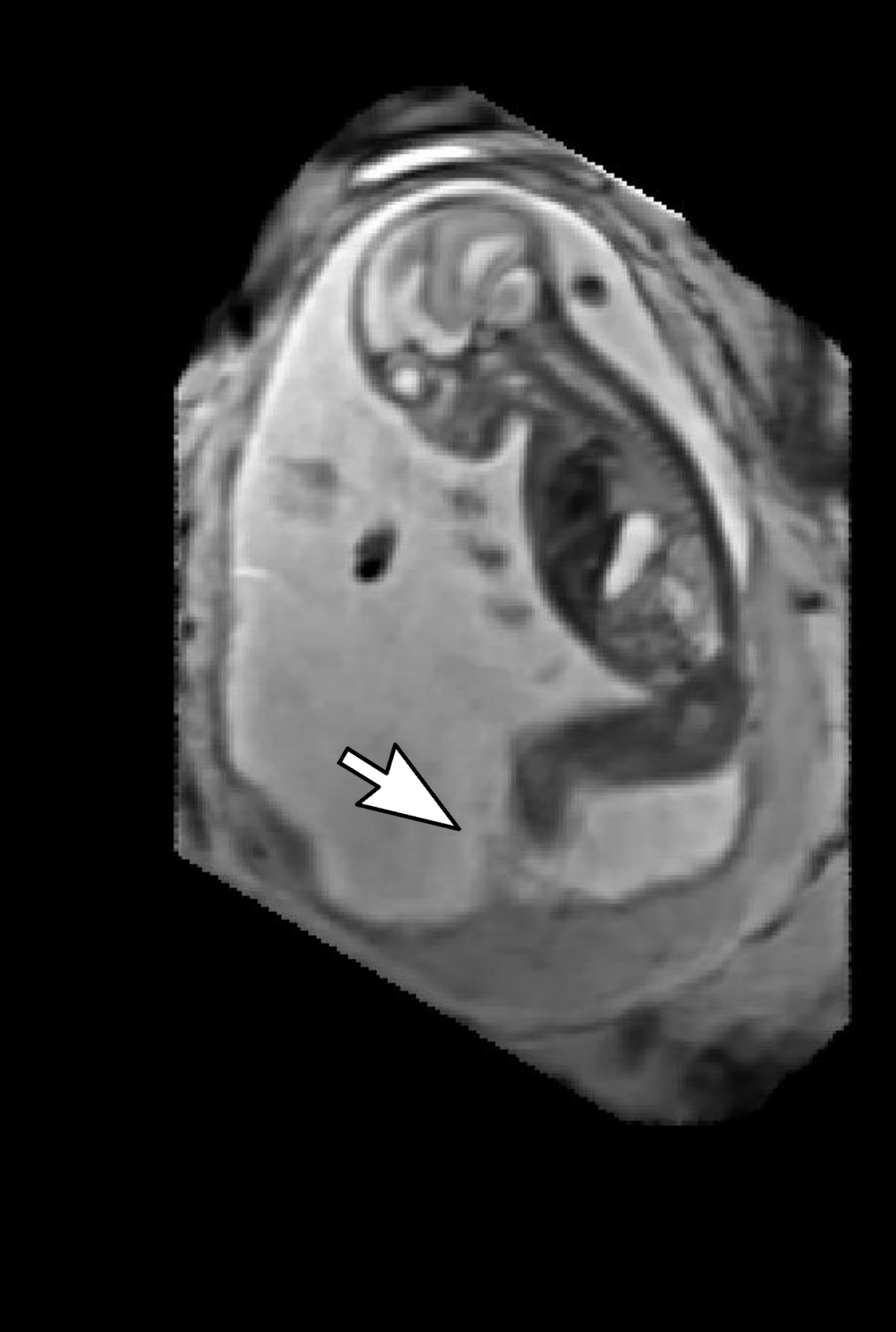}
	}
	\hfill	
	\subfloat{%
		\includegraphics[height=0.27\columnwidth,width=0.22\columnwidth,angle=0, trim=80 140 20 80,clip]{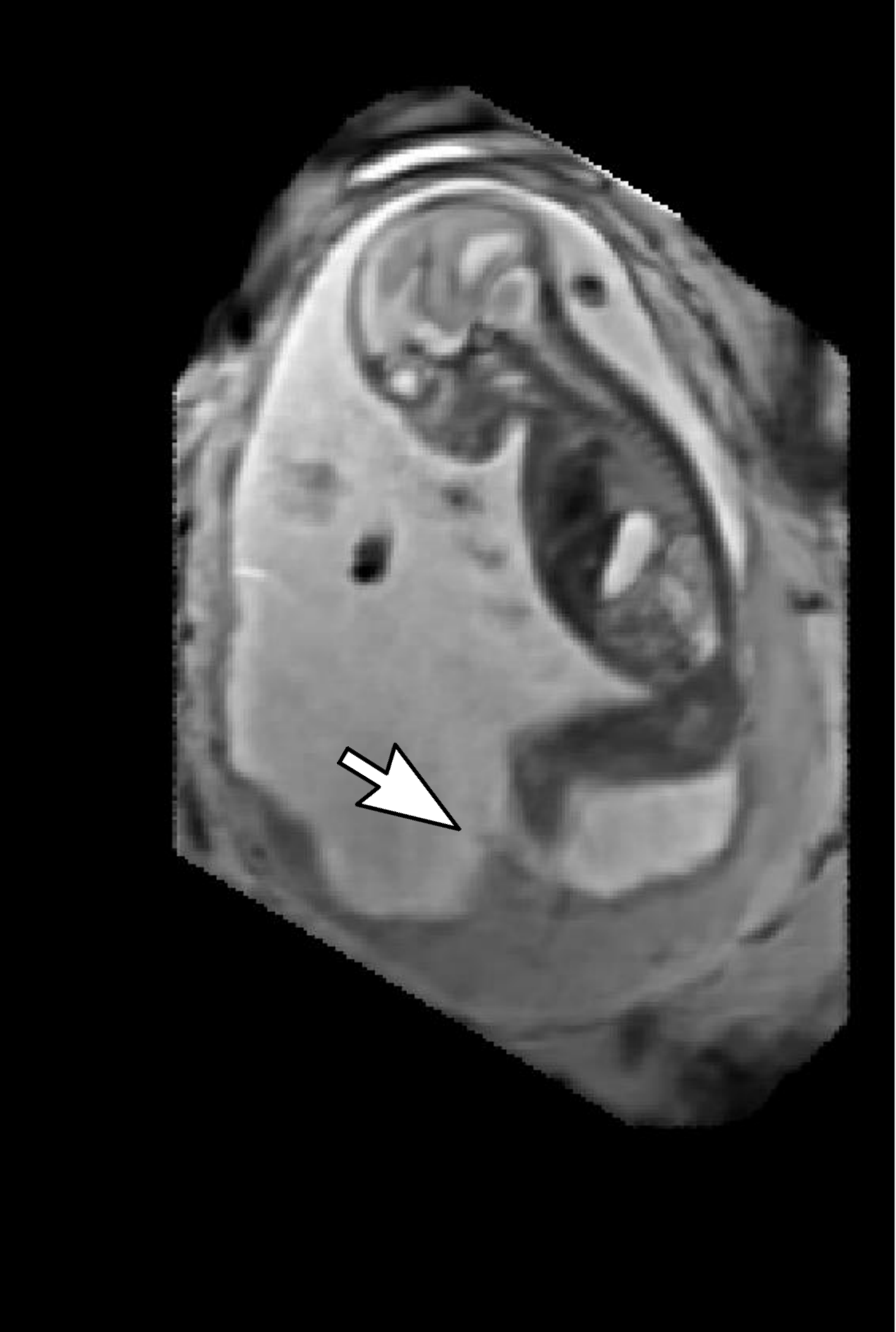}
	}
	\hfill
	\\
	\hfill
	\subfloat{
		\raisebox{0.35in}{\rotatebox[origin=t]{90}{(c) PVR}} 
	}
	\hfill
	\subfloat{%
		\includegraphics[width=0.22\columnwidth,angle=0, trim=0 0 0 0,clip]{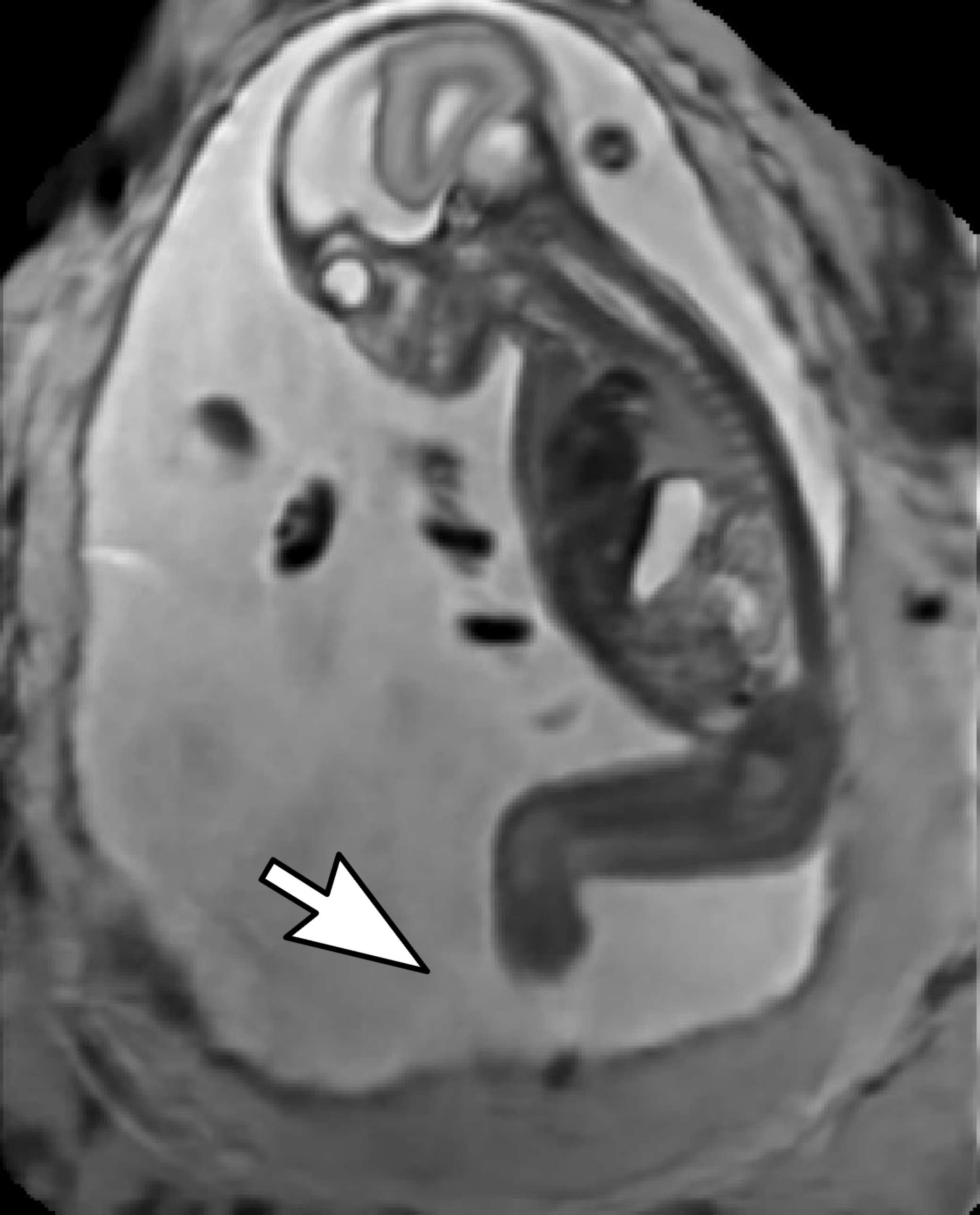}
	}
	\hfill
	\subfloat{%
		\includegraphics[width=0.22\columnwidth,angle=0, trim=0 0 0 0,clip]{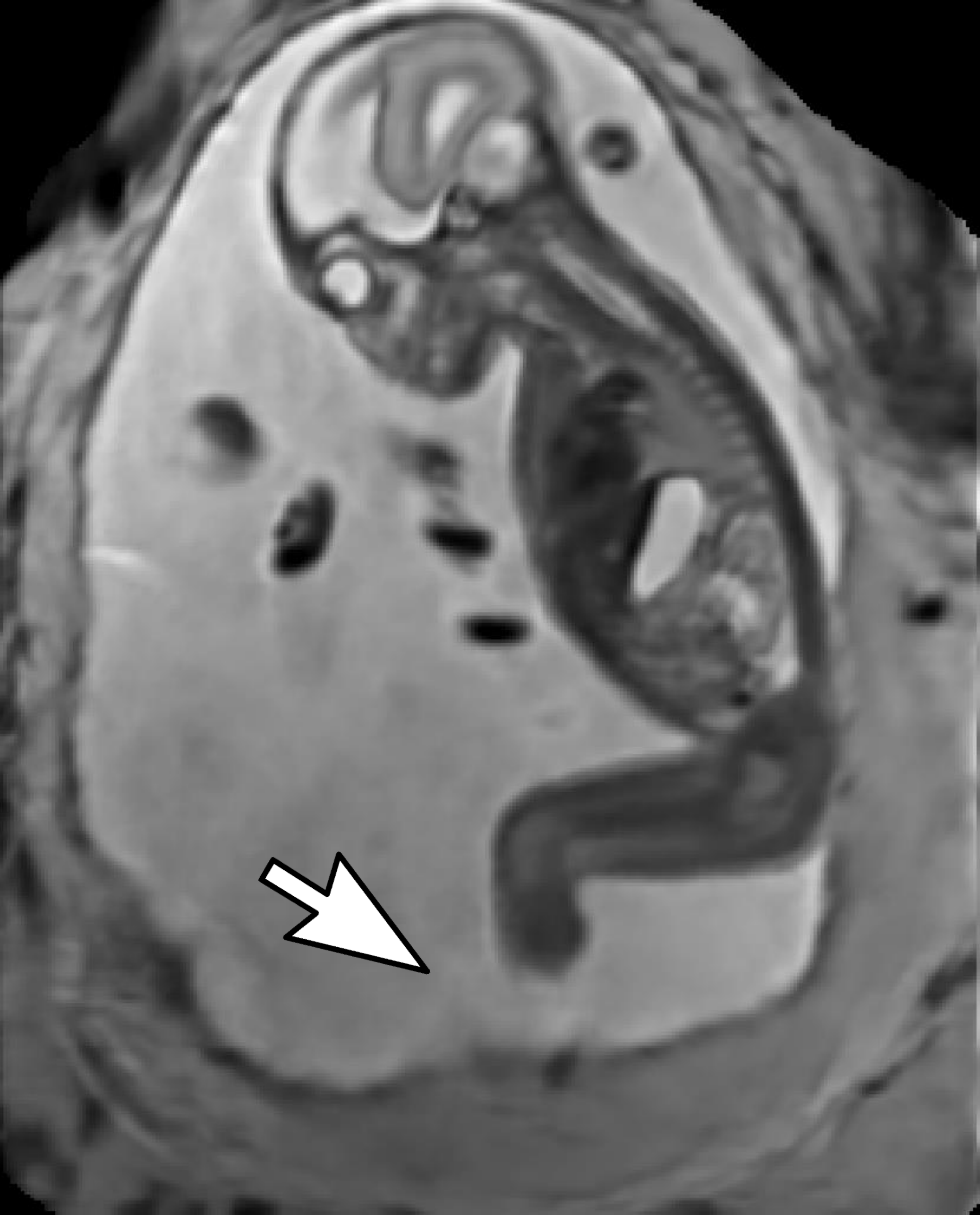}
	}
	\hfill
	\subfloat{%
		\includegraphics[width=0.22\columnwidth,angle=0, trim=0 0 0 0,clip]{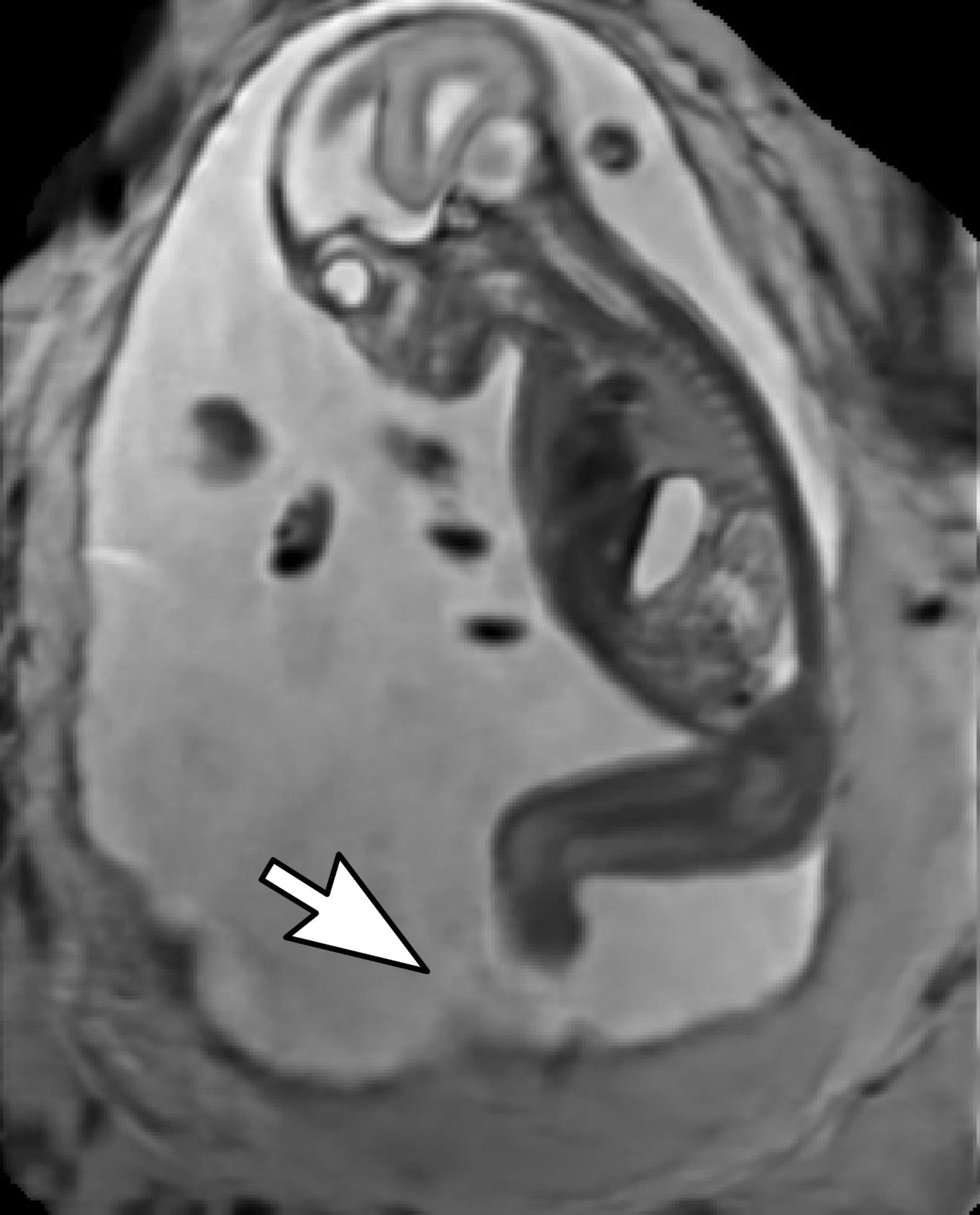}
	}
	\hfill	
	\subfloat{%
		\includegraphics[width=0.22\columnwidth,angle=0, trim=0 0 0 0,clip]{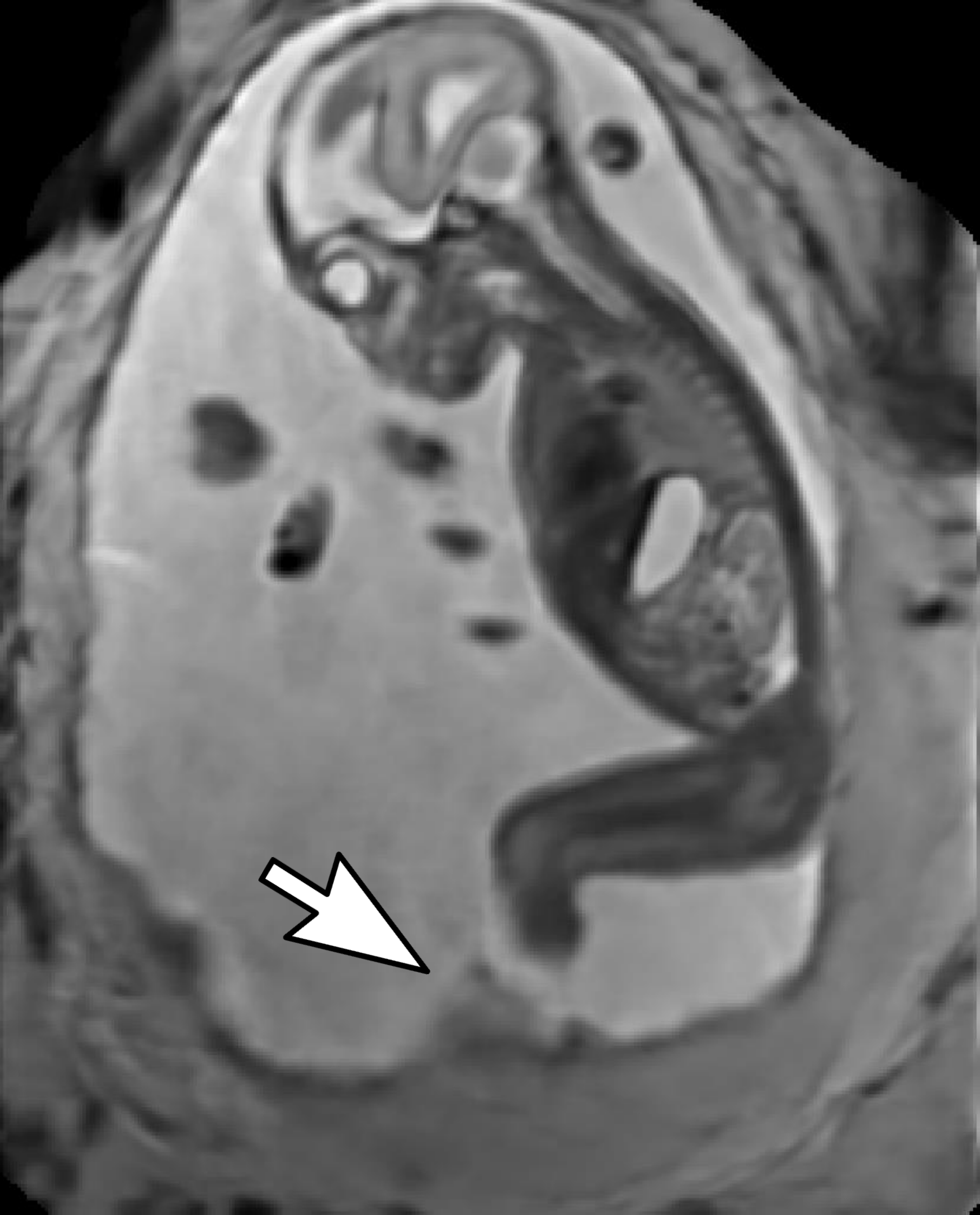}
	}
	\hfill\null
	\caption{Example reconstructions of consecutive MR scans of a moving fetus (kicking): input data (a) and corresponding cutting planes through a SVR-reconstructed (b) and PVR-reconstructed (c) volume. SVR produces blurry but readable results because of high data redundancy and outlier rejection through robust statistics. PVR with square patches of $a=32$ and $\omega=16$ appears visually superior.
	The arrow points at an area of substantial quality differences caused by independent rapid movements of the leg.}
\label{fig:fetal_motion}
\end{figure}

\textbf{Reconstruction of Fetal Organs:}
Exemplary PVR and SVR reconstructions under motion introduced by kicking of the fetus are shown in Fig. \ref{fig:fetal_motion}. PVR reconstruction results show an improved visual appearance and less blurring in the region with severe motion artifacts (arrow). An example of a challenging clinical case with a kidney malformation in one of twin fetuses, is shown in Fig. \ref{fig:twins}. Our clinical partners confirmed that such complications are easier to examine and to quantify after PVR-based reconstruction.

Comparative experiments of PVR variants were carried out on 32 fetal MR scans at gestational ages of approximately 20 weeks, presenting with challenging image corruption. Tab. \ref{table:evaluation_pvr} (a) \& (b) show numerical results of evaluating individual stacks before reconstruction (baseline), and the final reconstructed image using square patches, superpixels and multi-scale variants of PVR. Statistical testing between baseline and PVR variants was carried out using paired T-Tests and differences between using fixed or multi-scale and using square patches or superpixels were assessed via Two-factor ANOVA with repeated measures. In Tab. \ref{table:evaluation_pvr} (a) \& (b) the names of PVR variants are marked in bold if statistically significant differences have been found during analysis, \emph{i.e.}, \textbf{FS} and \textbf{MS} and/or \textbf{Square Patches} and \textbf{Superpixel} pairs are bold if the results between them differ significantly.

\begin{table*}[hbtp]
\caption{Average (a) PSNR and (b) SSIM results ($N=32$) for the input stack (baseline) and PVR variants with fixed (FS) and multi-scale (MS) variants of square patches and superpixels. All mean differences of PVR against baseline are statistically significant (p \textless 0.05). Names of all statistically significantly different PVR variants are stated in \textbf{bold}.}
\label{table:evaluation_pvr}

  \begin{minipage}{.5\linewidth}
    \centering
    \label{table:fetal_pvr_psnr}
    \resizebox{1.0\textwidth}{!}{
    	\begin{tabular}{lcccccccc}

        \multicolumn{9}{c} {\textbf{(a) PSNR}} \\
      
      	\toprule[1pt]
        & \multicolumn{2}{c}{\emph{Brain}}        & \multicolumn{3}{c}{\emph{Placenta}}     & \multicolumn{3}{c}{\emph{Uterus}}     \\
          \emph{Baseline}                         & \multicolumn{2}{c}{$16.97\pm3.77$}      & \multicolumn{3}{c}{$19.95\pm4.40$}			  & \multicolumn{3}{c}{$19.29\pm3.82$}      \\
         & Square Patches & Superpixels &  & \textbf{Square Patches} & \textbf{Superpixels}  &  & Square Patches & Superpixels \\
    
        \cmidrule(lr){2-3}\cmidrule(lr){5-6}\cmidrule(l){8-9} 
              
          \textbf{FS} & $26.70\pm1.45$ & $26.60\pm1.58$  & 
          \textbf{FS} & $31.07\pm1.35$ & $31.00\pm1.46$  &
          \textbf{FS} & $27.17\pm1.43$ & $27.35\pm1.36$ \\
          \textbf{MS} & $27.03\pm1.38$ & $26.35\pm1.54$  &  
          \textbf{MS} & $30.85\pm1.50$ & $29.62\pm1.30$  &  
          \textbf{MS} & $26.87\pm1.31$ & $26.40\pm1.24$ \\
        \bottomrule[1pt]
      \end{tabular}
    }
  \end{minipage}
  \begin{minipage}{.5\linewidth}
    \centering
    \label{table:fetal_pvr_ssim}
    \resizebox{1.0\textwidth}{!}{
    	\begin{tabular}{lcccccccc}
      
        \multicolumn{9}{c} {\textbf{(b) SSIM}} \\

        \toprule[1pt]
        & \multicolumn{2}{c}{\emph{Brain}} & \multicolumn{3}{c}{\emph{Placenta}} & \multicolumn{3}{c}{\emph{Uterus}}\\
          \emph{Baseline} & 
          \multicolumn{2}{c}{~$0.00\pm0.01$} & \multicolumn{3}{c}{~$0.00\pm0.02$}	  & \multicolumn{3}{c}{~$0.01\pm0.03$}\\
    & \textbf{Square Patches} & \textbf{Superpixels} & & \textbf{Square Patches} & \textbf{Superpixels}  &   & \textbf{Square Patches} & \textbf{Superpixels} \\

    \cmidrule(lr){2-3}\cmidrule(lr){5-6}\cmidrule(l){8-9} 

          \textbf{FS} & ~$0.51\pm0.03$ & ~$0.51\pm0.03$ &  
          \textbf{FS} & ~$0.58\pm0.03$ & ~$0.58\pm0.03$ &
          \textbf{FS} & ~$0.53\pm0.03$ & ~$0.50\pm0.03$ \\
          \textbf{MS} & ~$0.48\pm0.04$ & ~$0.45\pm0.05$  &
          \textbf{MS} & ~$0.54\pm0.04$ & ~$0.49\pm0.05$  & 
          \textbf{MS} & ~$0.50\pm0.03$ & ~$0.47\pm0.04$ \\
        \bottomrule[1pt]
      \end{tabular}
    }
  \end{minipage}
\end{table*}

\begin{figure*}[!htbp]
\centering
	\null\hfill
	\subfloat[\label{subfig:dssim_brain_input}]{%
		\includegraphics[width=0.22\columnwidth,height=0.22\columnwidth,trim=110 0 110 0,clip]{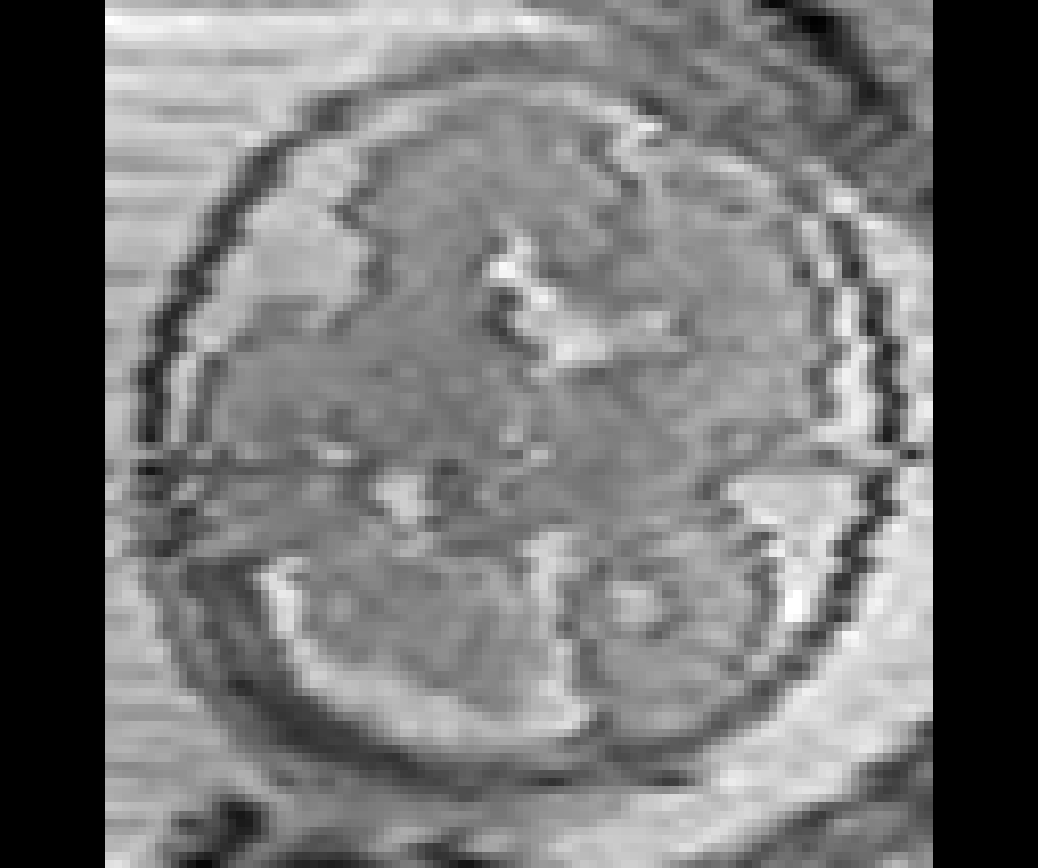}}
	\hfill
	\subfloat[\label{subfig:dssim_brain_recon}]{%
		\includegraphics[width=0.22\columnwidth,height=0.22\columnwidth,trim=110 0 110 0,clip]{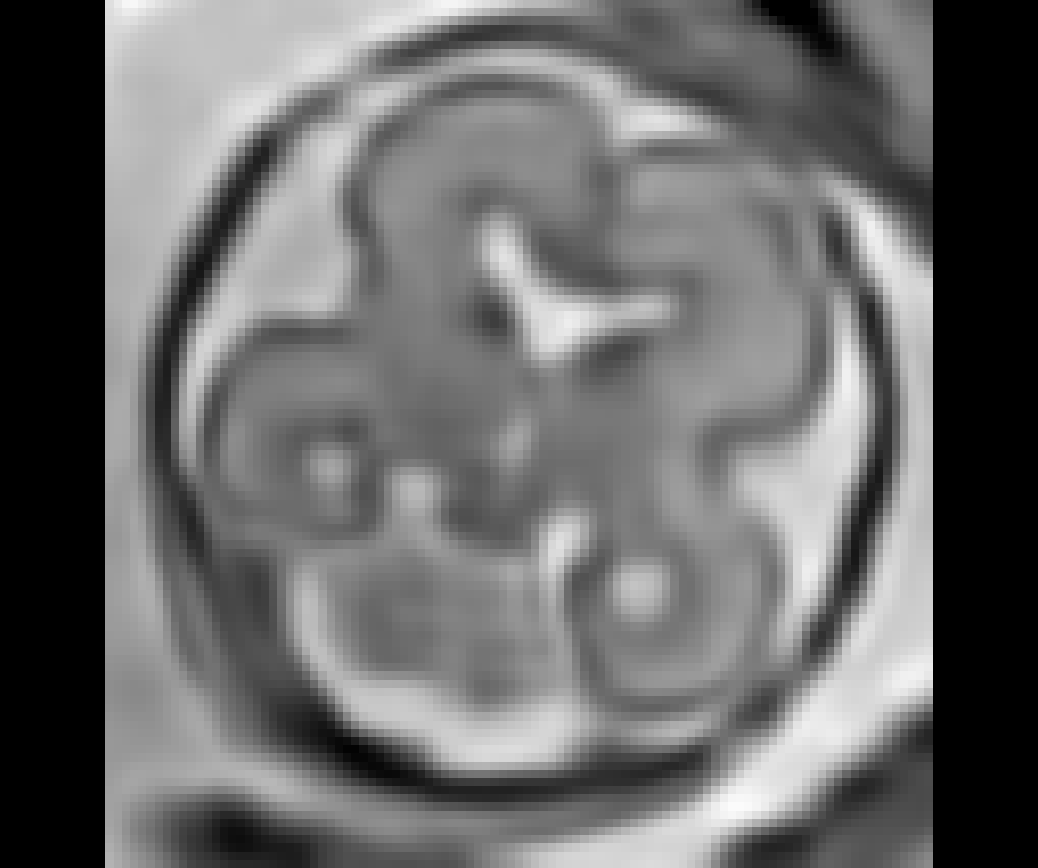}}	
	\hfill	
	\subfloat[\label{subfig:dssim_brain_baseline}]{%
		\includegraphics[width=0.22\columnwidth,height=0.22\columnwidth,trim=110 0 110 0,clip]{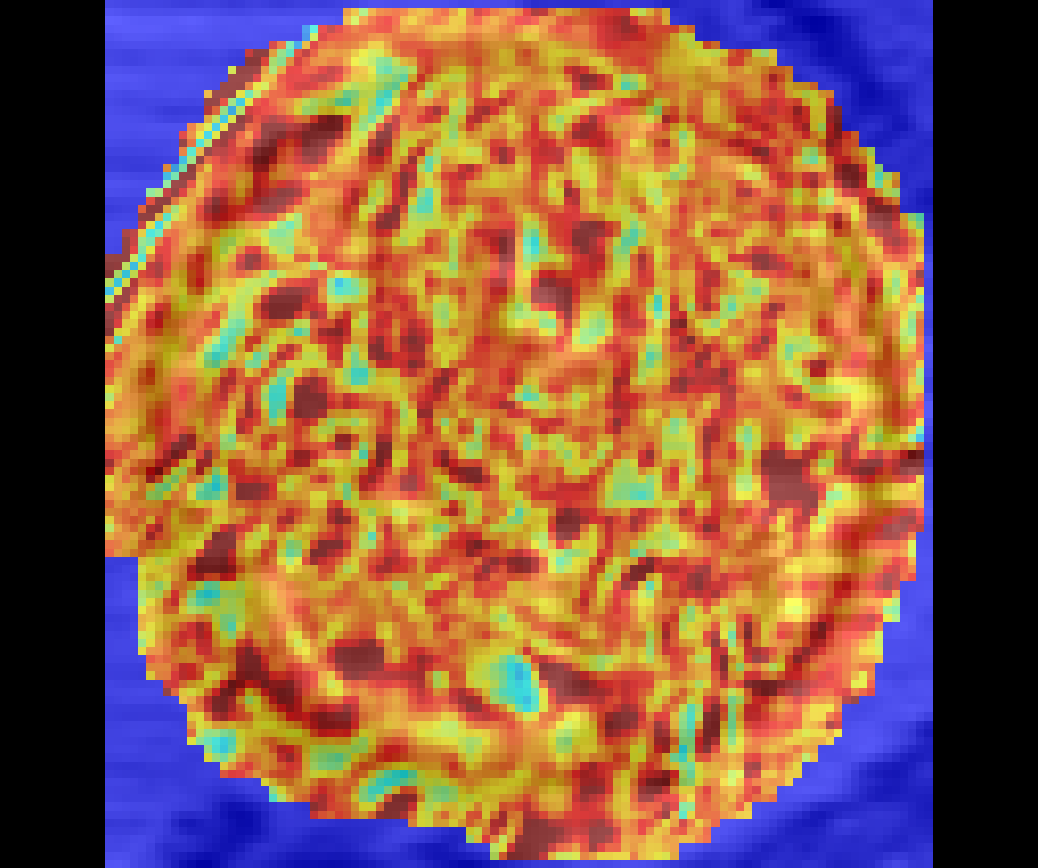}}
	\hfill	
	\subfloat[\label{subfig:dssim_brain_iter_4}]{%
		\includegraphics[width=0.22\columnwidth,height=0.22\columnwidth,trim=110 0 110 0,clip]{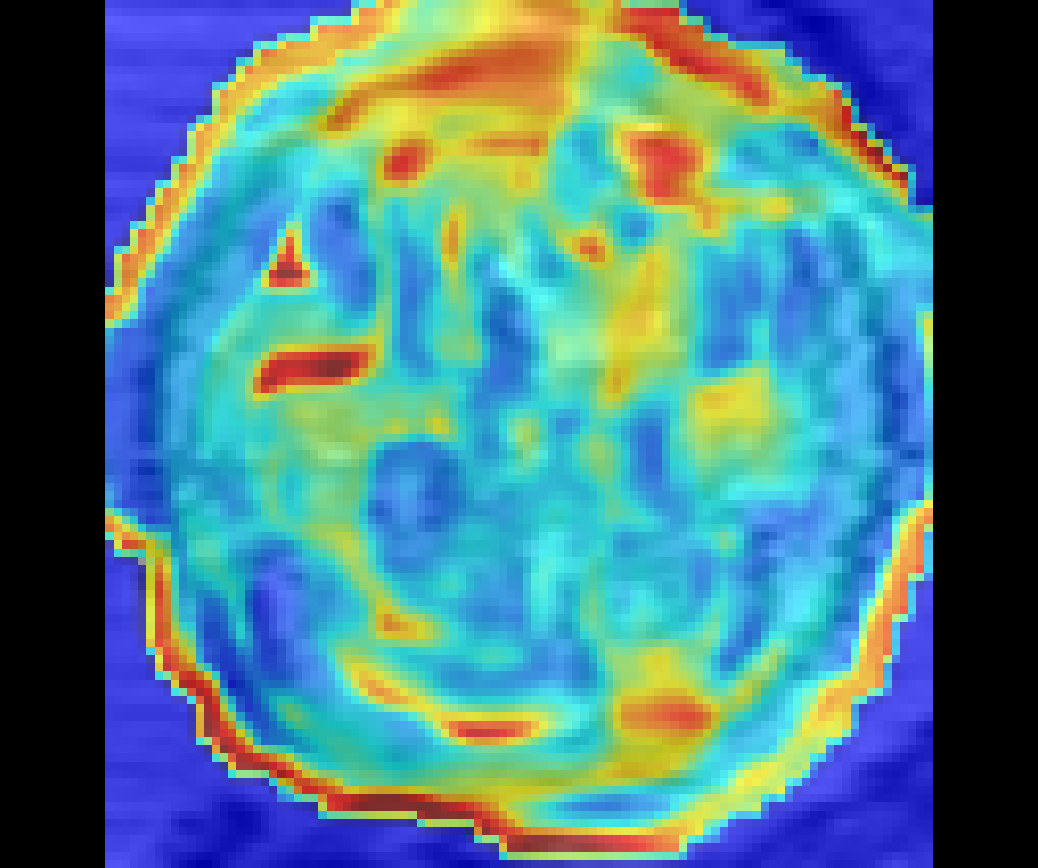}}	
	\hfill	
	\subfloat{%
		\includegraphics[height=0.22\columnwidth,trim=0 2 0 0,clip]{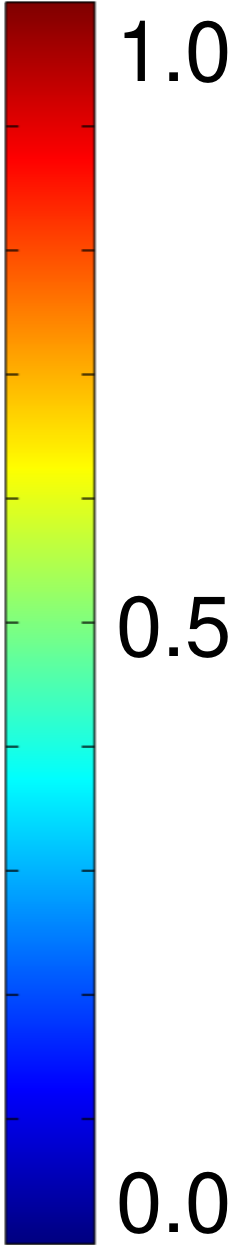}}
	\hfill\null
	\null\hfill
	\subfloat[\label{subfig:dssim_placenta_input}]{%
		\includegraphics[width=0.22\columnwidth,height=0.22\columnwidth,angle=90, trim=50 0 50 0,clip]{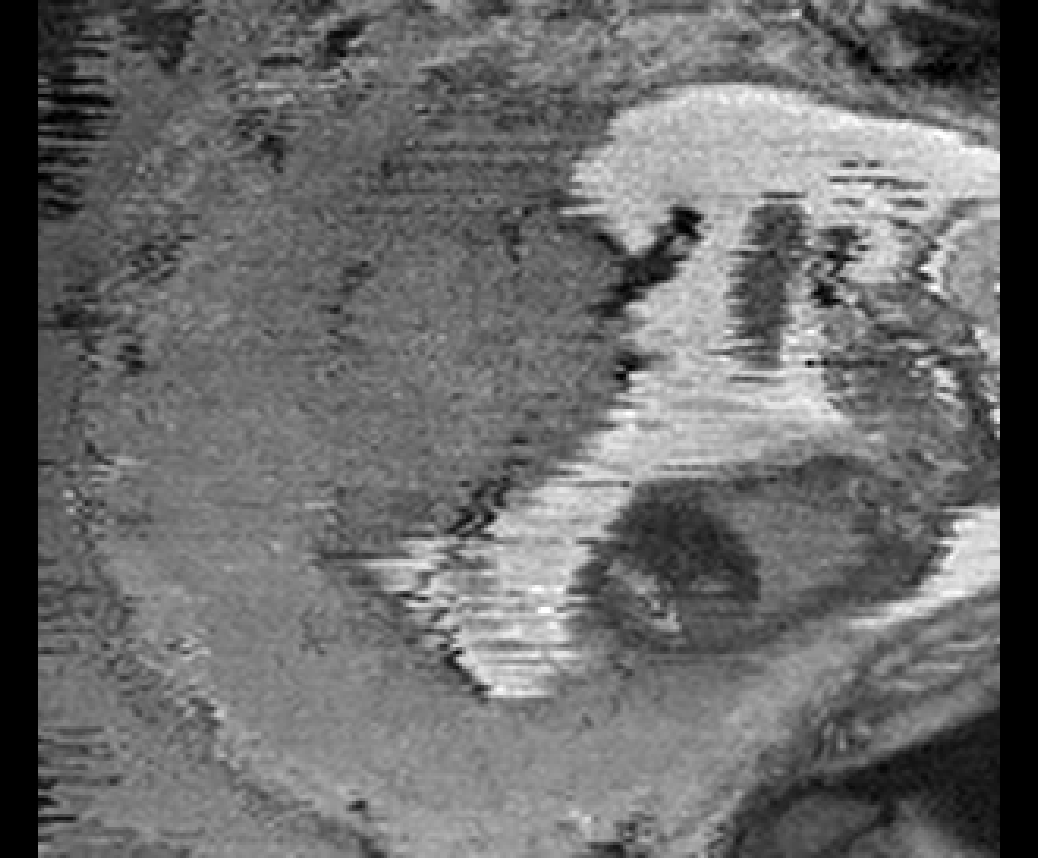}}
	\hfill
	\subfloat[\label{subfig:dssim_placenta_recon}]{%
		\includegraphics[width=0.22\columnwidth,height=0.22\columnwidth,angle=90, trim=50 0 50 0,clip]{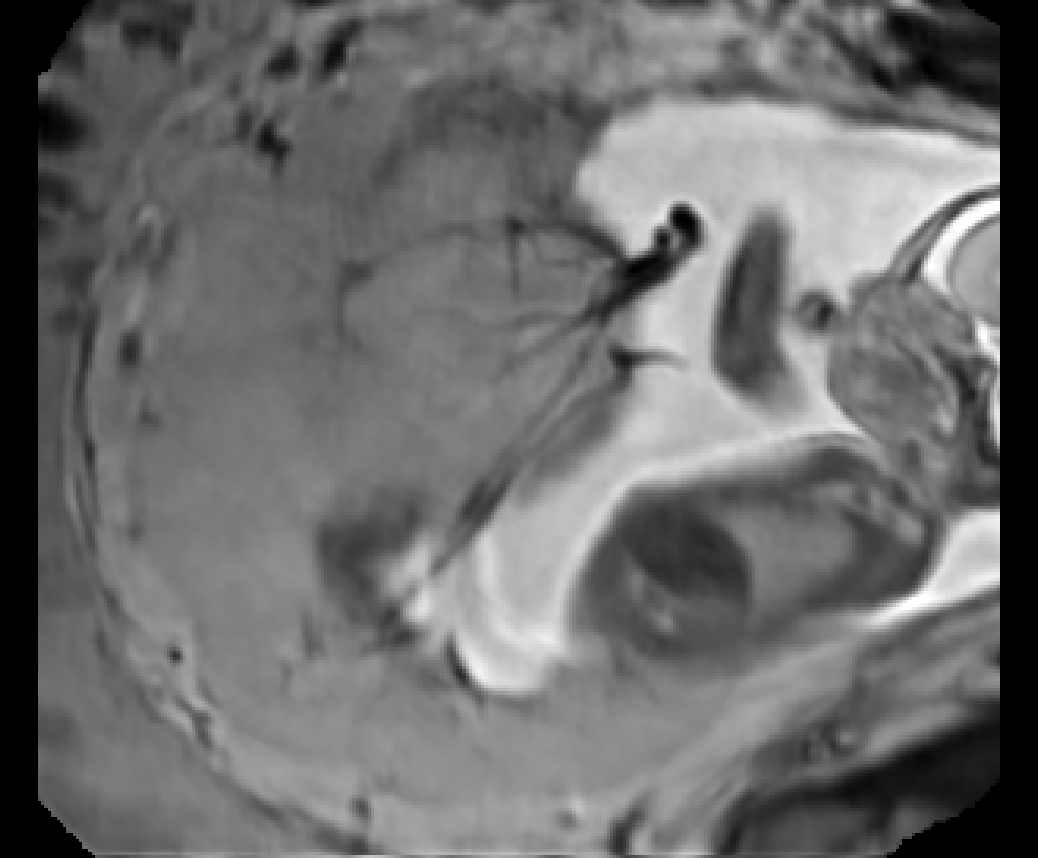}}
	\hfill
	\subfloat[\label{subfig:dssim_placenta_baseline}]{%
		\includegraphics[width=0.22\columnwidth,height=0.22\columnwidth,angle=90, trim=50 0 50 0,clip]{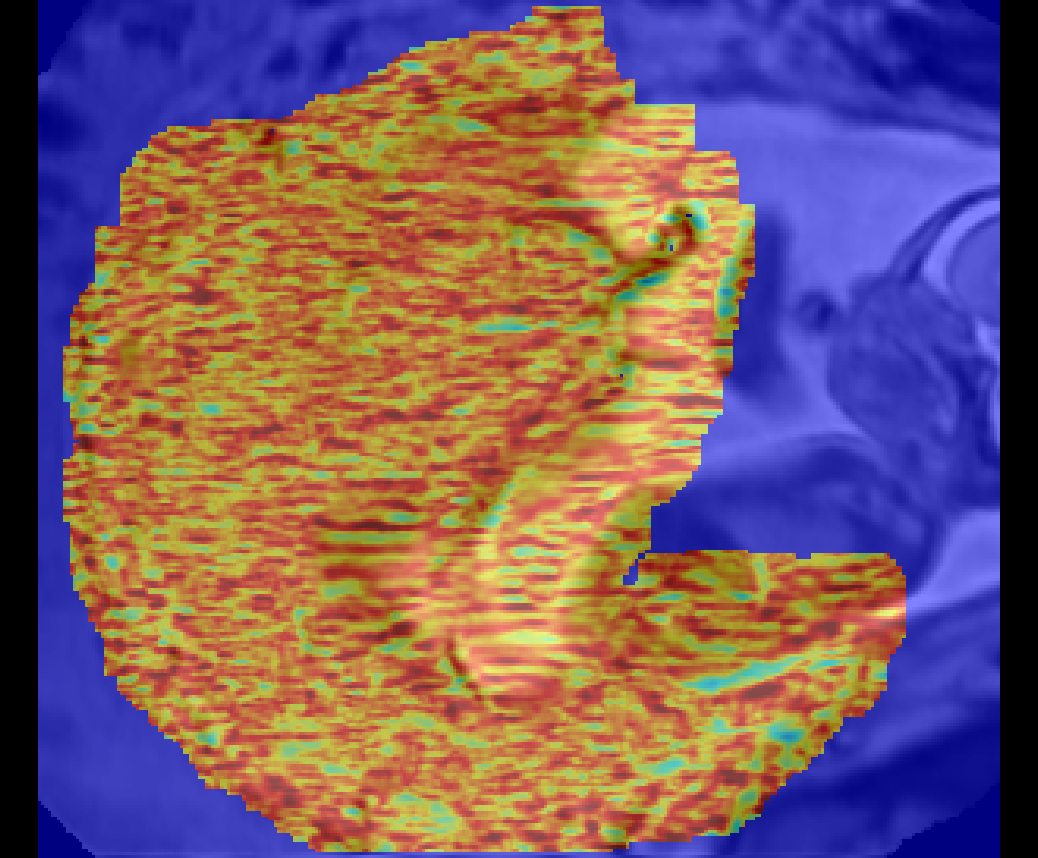}}
	\hfill	
	\subfloat[\label{subfig:dssim_placenta_iter_3}]{%
		\includegraphics[width=0.22\columnwidth,height=0.22\columnwidth,angle=90, trim=50 0 50 0,clip]{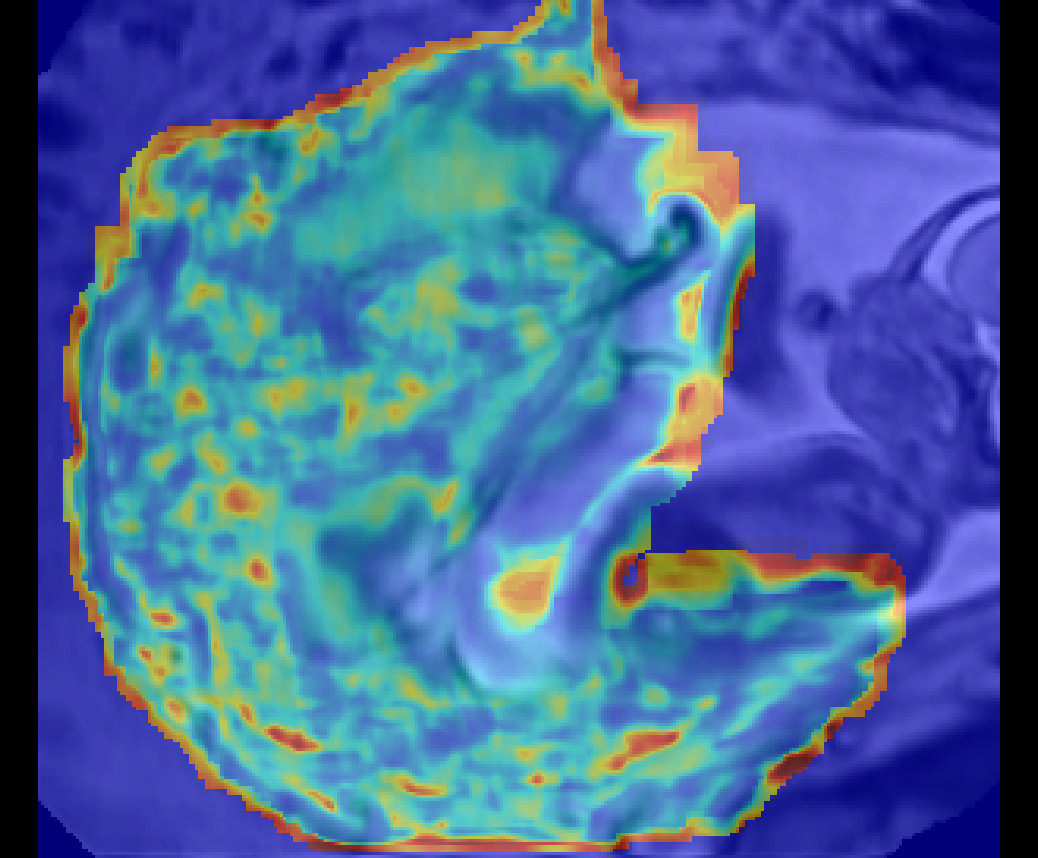}}
	\hfill
	\subfloat{%
		\includegraphics[height=0.22\columnwidth,trim=0 2 0 0,clip]{jet_bar.pdf}}
	\hfill\null

	\caption{A sample 2D cutting plane through a motion-corrupted fetal brain (a) and placenta (f), after PVR using square patches with $a=32$ and $\omega=16$ (b) and (g). The DSSIM heat map for a baseline before reconstruction (c) and (h), and after PVR (d) and (i). The average DSSIM of the fetal brain equals 0.497 (c) and 0.248 (d), while for the placenta equals to 0.491 (h) and 0.214 in (i).}
\label{fig:dssim}
\end{figure*}

The evaluation of the reconstruction quality of a whole 3D image into a single-valued metric may not properly reflect the performance differences, as it is based on averaging values of all the pixels of all the input stacks. Furthermore, Tab.~\ref{table:evaluation_pvr} indicates significant differences between variants of PVR but these differences have only minimal qualitative effect on reconstruction accuracy. 
Therefore, Fig.~\ref{fig:dssim} evaluates the reconstruction quality of PVR additionally using dissimilarity heat maps based on the measured DSSIM (see Sec. \ref{subsec:metrics}).
 This approach allows further qualitative evaluation and allows for uncertainty visualization of PVR reconstructions.

\begin{figure}[!htbp]
	\centering
	\subfloat[\label{subfig:dssim_brain_input}{Number of patches}]{%
		\includegraphics[width=1.0\columnwidth, trim=10 30 10 10,clip]{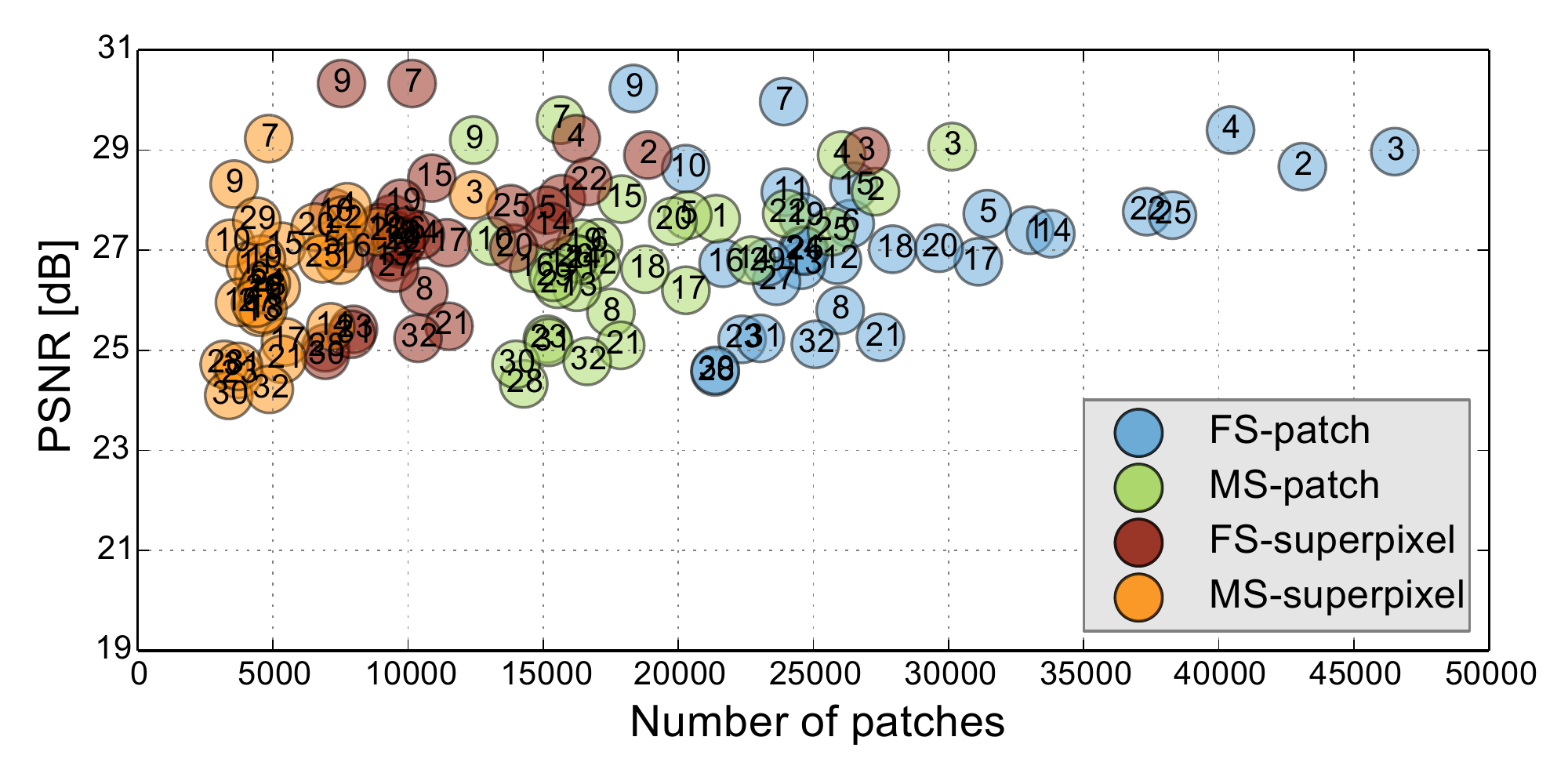}}
	\hfill
	\subfloat[\label{subfig:dssim_brain_recon}{Overhead pixels (\%)}]{%
		\includegraphics[width=1.0\columnwidth, trim=10 30 10 10,clip]{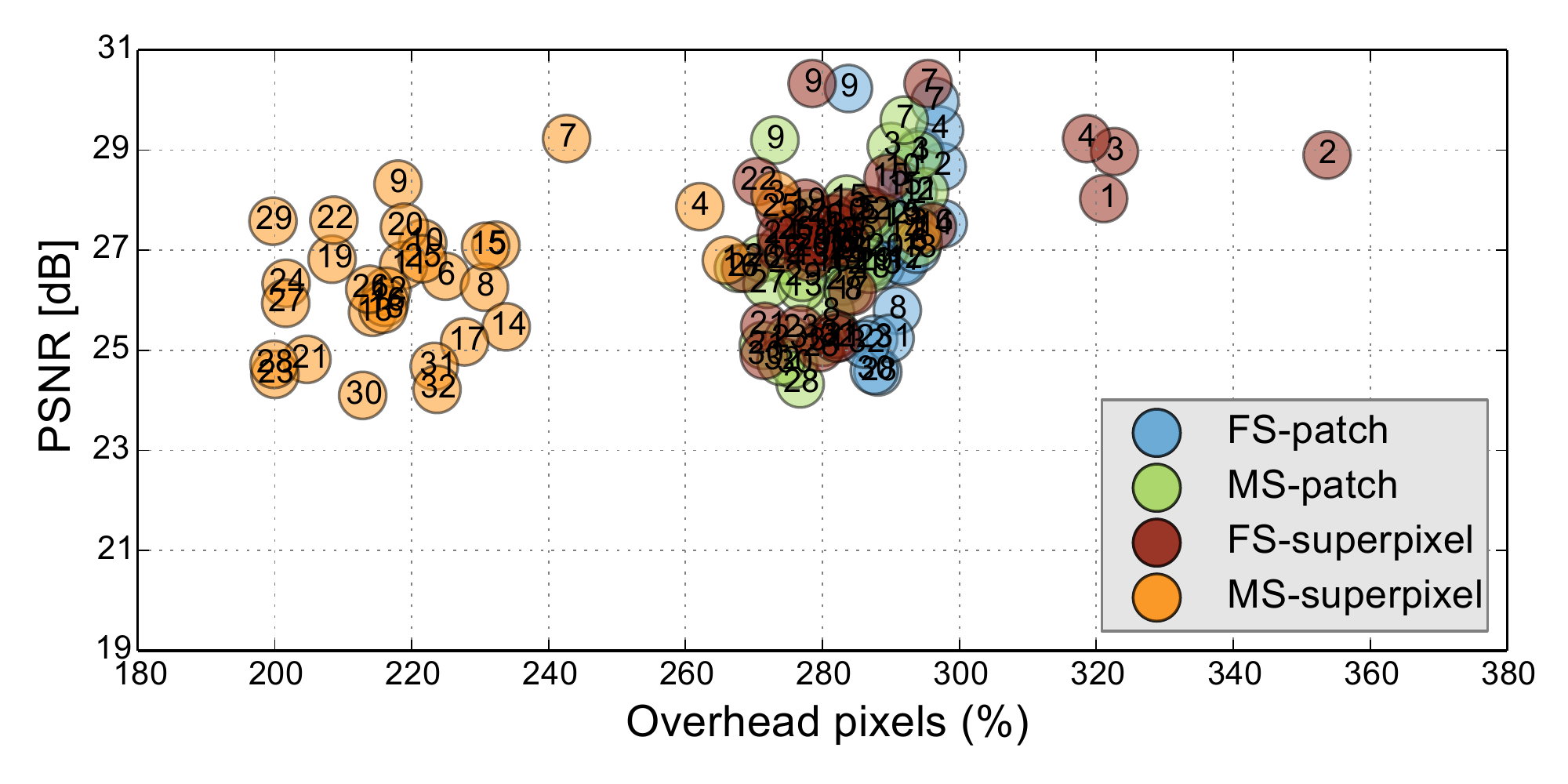}}	
	\caption{Number of generated patches (a) and necessary additional overhead pixels (\%) of the different PVR variants versus their reconstruction PSNR quality of the whole uterus (see Tab.~\ref{table:evaluation_pvr}-a). Optimal results are found in the upper left corner of the plots, \emph{i.e.}, high reconstruction quality and low computational overhead. The subject number is highlighted inside each circle marker. Multi-scale-superpixels (MS-superpixel) achieve similar reconstruction quality to fixed-size (FS-patch), multi-scale (MS-patch) square patches while clustering in the area of minimal computational overhead.}
	\label{fig:performance}
\end{figure}

\textbf{Performance Analysis:}
\label{subsec:perfotmance}
We further evaluate the computational performance of each PVR variant. Measuring the overall runtime is not meaningful because this would be highly machine specific and would include data transfer overhead and non optimized functions. The runtime varied between 2000--4000s on our testing machines, depending on the system configuration.
Instead we are analyzing the computational overhead introduced by PVR compared to SVR. The overhead can be measured by counting the number of processed patches and the number of additionally processed voxels. We compare these values to the achieved reconstruction quality in Fig.~\ref{fig:performance}. Multi-scale superpixels show significantly better performance than other PVR variants and introduce the minimum necessary overhead while gaining the same image quality than more na\"ive PVR variants. Multi-scale superpixels are potentially five times faster than other variants. 

\section{Discussion \& Conclusion}

We have introduced the concept of patch-to-volume reconstruction (PVR) in order to compensate non-rigid motion artifacts from fetal MRI scans without requiring  a defined region of interest. 
PVR splits the 3D input image into overlapping square patches and superpixels and employ automatic EM-based outlier rejection to find consistent data.

Our method is able to automatically reconstruct whole collections of motion corrupted stacks without the need of image segmentations and manual identification of rigid regions. We have shown that PVR can reconstruct the whole uterus, selected fetal organs, and secondary, non-rigidly moving pregnancy structures such as the placenta. 

PVR's reconstruction quality has been evaluated quantitatively and qualitatively on an adult phantom T2-weighted brain with synthetic non-rigid motion artifacts, as well as on the whole uterus from motion corrupted fetal MRI data including fetal brain, placenta and cases with multiple births.

PVR surpasses the state-of-the-art SVR method especially for considerable non-rigid deformations. We have evaluated different variants of PVR using fixed-size and multi-size square patches and fixed-size and multi-size superpixels. ANOVA analysis has shown significant differenced between these approaches for different areas of the uterus. However, evaluation of motion compensation methods is difficult especially due to the lack of ground truth in fetal MRI. Mapping the reconstruction quality of a whole 3D volume into a single-valued metric may not properly reflect qualitative differences, as it is based on averaging all measured values of all the input stacks.  Therefore, we have performed extensive qualitative analysis and present several examples and evaluation based on structural dissimilarity (DSSIM) heat maps. 

In addition to reconstruction and motion correction of the whole uterus, we  have also shown that our method works for multiple births cases with multiple fetuses sharing the same womb. These cases are more likely to have complications and to undergo MRI during pregnancy but would require extensive manual effort to be successfully reconstructed with state-of-the-art methods. 
 
Although our method is able to reconstruct the whole uterus automatically, small parts like limbs that move rapidly between the acquisition of individual slices are more difficult to recover. This is especially problematic for very young fetuses that have more space to move inside the womb. In cases of extreme limb movements (\textgreater2 cm between individual slices) PVR is not able to find structural consensus between overlapping patched and blurry image regions will be reconstructed. This is a general problem of automatic intensity-based optimization methods and different methods that are able to understand the semantic content of each patch will be required for future improvements.

PVR introduces a considerable computational overhead to the reconstruction stage of fetal MR image processing pipelines. We have evaluated the amount of necessary additional redundant information to give a general idea about the expected runtime of different PVR variants. Patches based on multi-scale superpixels are significantly more efficient than a na\"ive implementation of overlapping square patches, while maintaining a similar reconstruction accuracy. Quantitatively, square patches perform slightly better for the brain, which is most likely due to the rigid nature of the enclosing skull. Superpixel-based patches achieve better results for regions that are likely affected by non-rigid movements like the placenta and the whole uterus.



\section*{Acknowledgments}
We would like to thank volunteer subjects and the radiographers from St. Thomas Hospital London, especially Joanna Allsop and Matt Fox, for the image acquisitions. The data used in this research were collected subject to the informed consent of the participants. Access to the data will only be granted in line with that consent, subject to approval by the project ethics board and under a formal Data Sharing Agreement. For more details on our data sharing restrictions, visit \url{http://www.ifindproject.com/} and \url{http://www.developingconnectome.org/} to research group data sharing policy. We acknowledge MITK~\cite{wolf2005medical}, which we used to generate some of the figures. IRTK was used under creative commons public license from IXICO Ltd. We gratefully acknowledge the support of NVIDIA with the donation of a Tesla K40 GPU used for this research. We are also supported by the National Institute for Health Research (NIHR) Biomedical Research Centre based at Guy's and St Thomas' NHS Foundation Trust and King's College London. The views expressed are those of the author(s) and not necessarily those of the NHS, the NIHR or the Department of Health. Furthermore, this work was supported by Wellcome Trust and EPSRC IEH award iFIND [102431], ERC dHCP (FP/2007-2013 319456), and the EPSRC award EP/N024494/1. A. Alansary is supported by the Imperial College President's PhD Scholarship.



\ifCLASSOPTIONcaptionsoff
\newpage
\fi



\bibliographystyle{IEEEtran}
\bibliography{IEEEabrv,TMI2015}
\end{document}